\documentclass{article}

\usepackage{microtype}
\usepackage{graphicx}
\usepackage{subfigure}
\usepackage{booktabs} 

\usepackage{hyperref}
\usepackage[frozencache,cachedir=.]{minted}

\usepackage[accepted]{icml2025}

\usepackage{amsmath}
\usepackage{amssymb}
\usepackage{mathtools}
\usepackage{amsthm}
\usepackage{arydshln}

\usepackage[capitalize,noabbrev]{cleveref}

\theoremstyle{plain}
\newtheorem{theorem}{Theorem}[section]

\theoremstyle{definition}

\theoremstyle{remark}

\makeatletter
\newenvironment{smalleralign}[1][\small]
 {\par\nopagebreak\leavevmode\vspace*{-\baselineskip}%
  \skip0=\abovedisplayskip
  #1%
  \def\maketag@@@##1{\hbox{\m@th\normalfont\normalsize##1}}%
  \abovedisplayskip=\skip0
  \align}
 {\endalign\ignorespacesafterend}
\makeatother

\usepackage[textsize=tiny]{todonotes}

\usepackage{color}
\usepackage{bm}
\usepackage{amsfonts}
\usepackage{blindtext}
\usepackage{listings}
\usepackage{mathtools}
\usepackage[normalem]{ulem}
\usepackage{multicol}
\usepackage{multirow}
\usepackage{pifont}

\usepackage{graphbox}
\usepackage{makecell}
\usepackage{colortbl}
\usepackage{wrapfig}
\usepackage{array}
\newcommand\Tstrut{\rule{0pt}{2.6ex}}         
\newcommand\Bstrut{\rule[-0.9ex]{0pt}{0pt}}   

\usepackage{soul}
\colorlet{llgray}{lightgray!40}
\sethlcolor{llgray}

\newcommand{\data}{\text{data}}
\newcommand{\skipp}{\text{skip}}
\newcommand{\out}{\text{out}}
\newcommand{\inn}{\text{in}}
\newcommand{\noise}{\text{noise}}

\newcommand{\tref}{\theta_{\text{ref}}}

\DeclarePairedDelimiterX{\infdivx}[2]{(}{)}{%
	#1\;\delimsize\|\;#2%
}

\newcommand{\kl}{D_{\mathrm{KL}}\infdivx}
\newcommand{\js}{D_{\mathrm{JS}}\infdivx}

\newcommand{\vect}[1]{\bm{#1}}

\newcommand{\x}{\xv}

\newcommand{\dm}{\mathrm{d}}

\newcommand{\E}{\mathbb{E}}
\newcommand{\R}{\mathbb{R}}

\newcommand{\epsilonv}{\vect\epsilon}

\newcommand{\gv}{\vect g}

\newcommand{\sv}{\vect s}

\newcommand{\xv}{\vect x}

\newcommand{\zv}{\vect z}

\newcommand{\Dv}{\vect D}

\newcommand{\Fv}{\vect F}

\newcommand{\Iv}{\vect I}

\newcommand{\Dc}{\mathcal D}

\newcommand{\Lc}{\mathcal L}

\newcommand{\Nc}{\mathcal N}

\icmltitlerunning{Direct Discriminative Optimization}

\begin{document}

\twocolumn[
\icmltitle{Direct Discriminative Optimization: Your Likelihood-Based \\ 
Visual Generative Model is Secretly a GAN Discriminator}




\icmlsetsymbol{equal}{*}

\begin{icmlauthorlist}

\icmlauthor{Kaiwen Zheng}{thu,nv}
\icmlauthor{Yongxin Chen}{nv}
\icmlauthor{Huayu Chen}{thu}
\icmlauthor{Guande He}{ut}
\icmlauthor{Ming-Yu Liu}{nv}
\icmlauthor{Jun Zhu}{thu}
\icmlauthor{Qinsheng Zhang}{nv}

\end{icmlauthorlist}

\centering

\textbf{\url{https://research.nvidia.com/labs/dir/ddo/}}

\icmlaffiliation{thu}{Tsinghua University}
\icmlaffiliation{ut}{The University of Texas at Austin}
\icmlaffiliation{nv}{NVIDIA}

\icmlcorrespondingauthor{Jun Zhu}{dcszj@tsinghua.edu.cn}

\icmlkeywords{Machine Learning, ICML}

\vskip 0.3in
]



\printAffiliationsAndNotice{} 

\begin{abstract}
While likelihood-based generative models, particularly diffusion and autoregressive models, have achieved remarkable fidelity in visual generation, the maximum likelihood estimation (MLE) objective, which minimizes the forward KL divergence, inherently suffers from a mode-covering tendency that limits the generation quality under limited model capacity. In this work, we propose Direct Discriminative Optimization (DDO) as a unified framework that integrates likelihood-based generative training and GAN-type discrimination to bypass this fundamental constraint by exploiting reverse KL and self-generated negative signals. Our key insight is to parameterize a discriminator implicitly using the likelihood ratio between a learnable target model and a fixed reference model, drawing parallels with the philosophy of Direct Preference Optimization (DPO). Unlike GANs, this parameterization eliminates the need for joint training of generator and discriminator networks, allowing for direct, efficient, and effective finetuning of a well-trained model to its full potential beyond the limits of MLE. DDO can be performed iteratively in a self-play manner for progressive model refinement, with each round requiring less than 1\% of pretraining epochs. Our experiments demonstrate the effectiveness of DDO by significantly advancing the previous SOTA diffusion model EDM, reducing FID scores from 1.79/1.58/1.96 to new records of 1.30/0.97/1.26 on CIFAR-10/ImageNet-64/ImageNet 512$\times$512 datasets without any guidance mechanisms, and by consistently improving both guidance-free and CFG-enhanced FIDs of visual autoregressive models on ImageNet 256$\times$256.
\end{abstract}

\begin{figure}[ht!]

	\centering
	\begin{minipage}{1.0\linewidth}
		\centering
			\includegraphics[width=\linewidth]{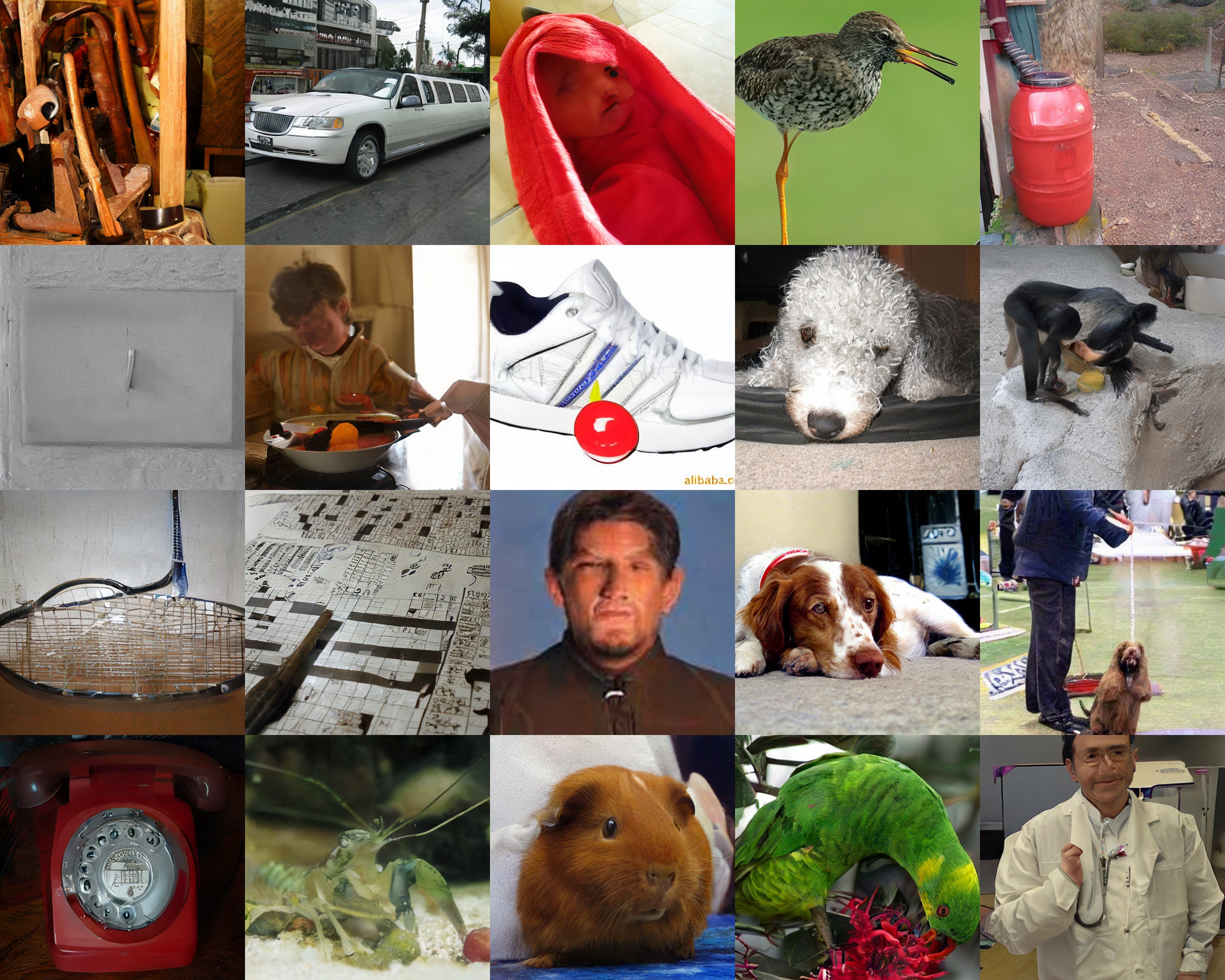}\\
\small{(a) EDM2-L~\citep{karras2024analyzing} (FID 1.96)}
	\end{minipage}
	\begin{minipage}{1.0\linewidth}
	\centering
	\includegraphics[width=\linewidth]{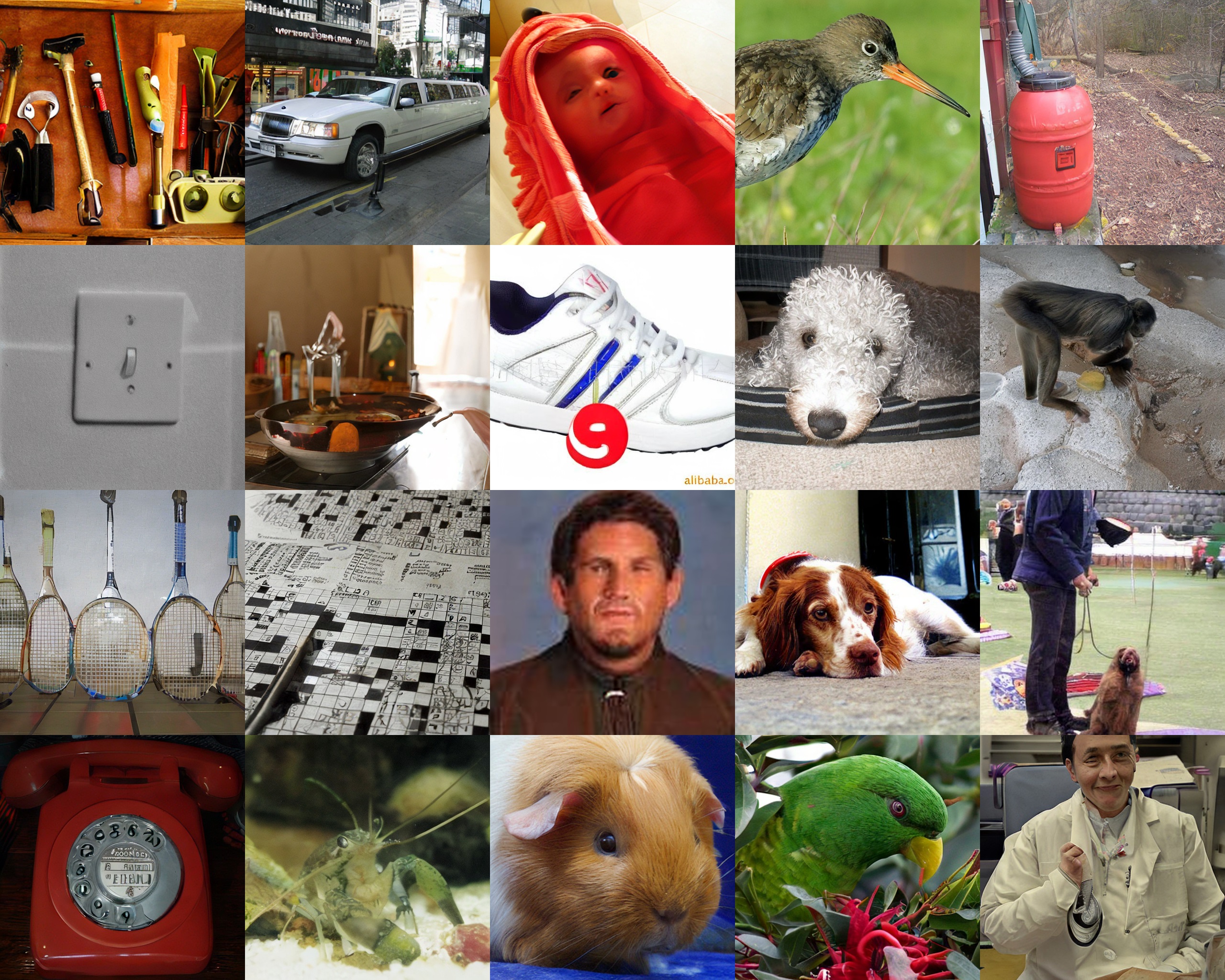}\\
\small{(b) EDM2-L + DDO (Ours, FID \textbf{1.26})}
\end{minipage}
   \vspace{-.15in}
	\caption{\label{fig:img512} 
Samples on ImageNet 512$\times$512, \textit{without any guidance}.}
	\vspace{-.25in}
\end{figure}

\section{Introduction}
Modeling the distribution of high-dimensional data is a fundamental challenge in machine learning~\citep{bishop2006pattern,Goodfellow2016}. Recent years have witnessed the domination of diffusion~\citep{ho2020denoising,song2020score} and autoregressive~\citep{van2016pixel} paradigms in generative modeling of continuous data and discrete data. They have achieved both theoretical and empirical success in visual tasks including image and video synthesis~\citep{dhariwal2021diffusion,esser2021taming,ramesh2021zero,karras2022elucidating,ho2022imagen,rombach2022high,balaji2022ediff,gupta2023photorealistic,esser2024scaling,brooks2024video,bao2024vidu,tian2024visual}, forming the cornerstone of large-scale generation systems in the era of AI-generated content.

\begin{figure}[t]

	\centering
	\begin{minipage}{.48\linewidth}
		\centering
			\includegraphics[width=.9\linewidth]{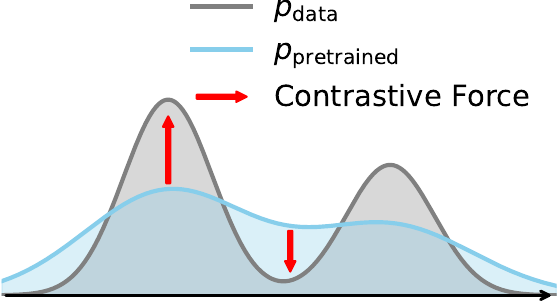}\\
\small{(a)}
	\end{minipage}
	\begin{minipage}{.48\linewidth}
	\centering
	\includegraphics[width=.9\linewidth]{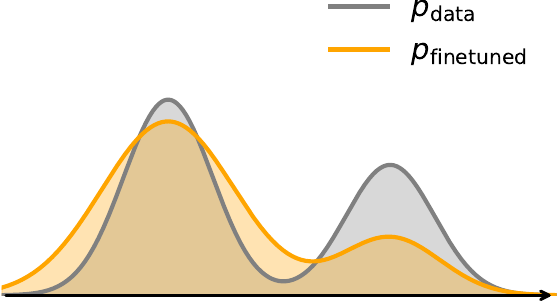}\\
\small{(b)}
\end{minipage}
   \vspace{-.05in}
	\caption{\label{fig:toy} Toy example illustrating DDO. (a) Models pretrained via maximum likelihood estimation (MLE) exhibit dispersed density, while DDO imposes contrastive forces toward the data distribution. (b) The finetuned model concentrates better on the main mode.}
	\vspace{-.15in}
\end{figure}

Diffusion and autoregressive models are representatives of likelihood-based generative models. Compared to Generative Adversarial Networks (GANs)~\citep{goodfellow2014generative} which often face unstable training and mode collapse issues, these models are distinguished by their stability, scalability, and generalizability. Besides, their iterative generation process imposes fewer constraints on network Lipschitzness, potentially facilitating superior generative capability over GAN's single-step generation. Likelihood-based generative models aim to learn the underlying data distribution $p_\data$ by maximizing the likelihood of the observed data under a parameterized probabilistic model $p_\theta$, which is equivalent to minimizing the forward Kullback–Leibler (KL) divergence:
\begin{align*}
    \max_{\theta} \E_{p_\data(\x)}\left[\log p_\theta(\x)\right]\Longleftrightarrow \min_{\theta} \kl{p_\data}{p_\theta}
\end{align*}
However, this maximum likelihood estimation (MLE) objective entails inherent limitations. Forward KL is known to prioritize ``mode-covering" and imposes extreme penalties if the model severely underestimates the likelihood of any training sample~\citep{karras2024guiding}. Under limited model capacity, this property forces the learned density to spread out excessively (Figure~\ref{fig:toy}(a)), potentially leading to blurry samples—a phenomenon commonly observed in Variational Autoencoders (VAEs)~\citep{kingma2013auto} and in likelihood training of diffusion models~\citep{song2021maximum,kingma2021variational,lu2022maximum,zheng2023improved}. Consequently, these models often rely heavily on guidance methods~\citep{ho2021classifier,kim2023refining,karras2024guiding} to steer the samples away from unlikely low-probability regions and toward the core of the data manifold in order to improve overall generation quality. In contrast, GANs, which are theoretically grounded in Jensen–Shannon (JS) divergence or Wasserstein distance~\citep{arjovsky2017wasserstein}, tend to produce sharper and more realistic samples.

To bypass MLE's mode-covering nature, we aim to leverage GAN-type loss to discriminate between the model and data distributions and produce contrastive forces that guide the model. However, typical GANs require parameterizing extra discriminator networks and alternating optimization, creating engineering complications. Applying GAN-type training trivially to diffusion or autoregressive models is especially inefficient due to their iterative sampling processes.

In this work, we introduce Direct Discriminative Optimization (DDO), a framework that bridges likelihood-based generative models with GANs to push their performance beyond the limits of MLE. Our key insight is to implicitly parameterize the discriminator using the likelihood ratio between a learnable target model and a fixed reference model, both initialized from the pretrained model. This parameterization, inspired by Direct Preference Optimization (DPO)~\citep{rafailov2024direct}, offers theoretical guarantees of optimality, divergence bounds, and connections to guidance methods. It also enables direct finetuning of the pretrained model without altering the network architecture or inference protocol and supports iterative refinement via multi-round self-play.

DDO achieves significant performance gains for both diffusion and autoregressive models sufficiently pretrained on standard image benchmarks. By finetuning state-of-the-art diffusion models EDM~\citep{karras2022elucidating} and EDM2~\citep{karras2024analyzing}, we achieve unprecedented guidance-free FID scores of 1.30/0.97/1.26 on CIFAR-10/ImageNet-64/ImageNet 512$\times$512 datasets. Finetuning the visual autoregressive model VAR~\citep{tian2024visual} on ImageNet 256$\times$256 reduces the FID from 1.92 to 1.73 while removing sampling tricks. Notably, even without classifier-free guidance (CFG)~\citep{ho2021classifier}, the finetuned model achieves an FID of 1.79, surpassing the CFG-enhanced pretrained model while cutting the inference cost by half.
\section{Background}
\subsection{Likelihood-Based Generative Models}
\label{sec:bg1}
Likelihood-based generative models parameterize a probability distribution $p_\theta$ to learn the data distribution $p_\data$, enabling explicit likelihood evaluation and density estimation. Among them, diffusion and autoregressive models are two prominent types that excel in visual generation.

Autoregressive (AR) models~\citep{van2016pixel,brown2020language} learn discrete\footnote{AR can also be adapted to model continuous data~\citep{tschannen2024givt,li2024autoregressive}.} data distributions via the next-token prediction mechanism: 
\begin{equation}
    \label{eq:ar_logp}
    \log p_\theta(\x)=\sum_{n=1}^d\log p_\theta(x^{(n)}|\x^{(<n)})
\end{equation}
where $d$ denotes the data dimension (sequence length). It factorizes the joint distribution into a product of conditional probabilities, allowing exact likelihood computation. Each $p_\theta(\cdot|\x^{(<n)})$ is parameterized via a Softmax operation over the model’s output logits and optimized using cross-entropy loss against the ground-truth token. In visual autoregressive modeling, images are first quantized to discrete tokens within a compact latent space using autoencoders~\citep{van2017neural,esser2021taming}.

Diffusion models~\citep{ho2020denoising,song2020score} learn continuous data distributions by gradually perturbing clean data $\x_0\sim p_\data$ with Gaussian noise, which generates a trajectory $\{\x_t\}_{t=0}^T$, and then learning to reverse this process. The forward and backward dynamics can be formulated as either stochastic or ordinary differential equations (SDEs or ODEs)~\citep{song2020score}. 
The forward process follows a closed-form transition kernel $q_{t|0}(\x_t|\x_0)=\Nc(\alpha_t\x_0,\sigma_t^2\Iv)$ with predefined noise schedule $\alpha_t,\sigma_t$, enabling reparameterization as $\x_t=\alpha_t\x_0+\sigma_t\epsilonv,\epsilonv\sim\Nc(\vect 0,\Iv)$. The model is typically parameterized as a noise prediction network $\epsilonv_\theta(\x_t,t)$ trained to estimate $\epsilonv$ via mean squared error (MSE) regression, which forms an evidence (or variational) lower bound (ELBO) on the likelihood~\citep{song2021maximum}:
\begin{equation}
    \label{eq:diffusion_logp}
    \log p_\theta(\x_0)\geq C-\E_{t\sim p(t),\epsilonv\sim\Nc(\vect0,\Iv)}\left[w(t)\|\epsilonv_\theta(\x_t,t)-\epsilonv\|_2^2\right]
\end{equation}
where $\x_t=\alpha_t\x+\sigma_t\epsilonv$, $C$ is a constant irrelevant to $\theta$, and $p(t),w(t)$ are certain time distribution and weighting function. The ELBO provides a reasonable likelihood approximation compared to the exact but cumbersome instantaneous change-of-variable formula in neural ODEs~\citep{chen2018neural}. Moreover, while the likelihood bound is tight only for specific $p(t), w(t)$, alternative choices share the same optimum and can serve as surrogate objectives~\citep{kingma2024understanding}.

From the perspective of score matching~\citep{song2020score}, the optimal noise predictor is linked to the \textit{score function} $\sv^*(\x_t,t)\coloneqq\nabla_{\x_t}\log q_t(\x_t)$ by $\epsilonv^*(\x_t)=-\sigma_t\sv^*(\x_t,t)$, where $q_t$ denotes the marginal distribution at time $t$ in the forward process. Due to the properties of MSE, the network can be parameterized in alternative yet theoretically equivalent forms, such as a velocity predictor~\citep{salimans2022progressive,zheng2023improved} that estimates the tangent of the diffusion trajectory, commonly known as flow matching~\citep{lipman2022flow}. In our experiments, we adopt the more generalized $F$-parameterization introduced in EDM~\citep{karras2022elucidating} (detailed in Appendix~\ref{appendix:exp-details}).
\subsection{GANs}
\label{sec:bg2}
GANs~\citep{goodfellow2014generative} do not explicitly model the likelihood $p_\theta$ but instead directly optimize the data generation process through adversarial training. Specifically, the optimization involves an adversarial interplay between a generator network $\gv_\theta:\R^{d_z}\mapsto\R^d$ that maps latent variables $\zv\in\R^{d_z}\sim p(\zv)$ (typically Gaussian noise) into synthetic samples, and a discriminator network $d_\phi:\R^d\mapsto[0,1]$ that classifies samples as real or fake:
\begin{equation}
\min_\theta\max_\phi\E_{p_\data(\x)}\left[\log d_\phi(\x)\right]+\E_{p_\theta(\x)}\left[\log(1-d_\phi(\x))\right].
\end{equation}
Here $p_\theta(\x)$ is the generator distribution, whose exact density is intractable but can be easily sampled from via $\x=\gv_\theta(\zv),\zv\sim p(\zv)$. In the inner loop, the discriminator is optimized using binary cross-entropy loss (also known as noise contrastive estimation (NCE)~\citep{gutmann2010noise}), and its optimal solution can be derived as:
\begin{equation}
\label{eq:optimal-discriminator}
    d^*(\x)=\frac{p_\data(\x)}{p_\data(\x)+ p_\theta(\x)}
\end{equation}
under which the minimax game becomes
\begin{equation} 
\min_\theta 2\js{p_\data}{p_\theta} - 2 \log 2
\end{equation}
where $\js{p}{q}=\frac{1}{2}\kl{p}{\frac{p+q}{2}}+\frac{1}{2}\kl{q}{\frac{p+q}{2}}$ is the Jensen–Shannon (JS) divergence. This theoretically ensures that the optimal generator distribution matches the data distribution. However, in practice, training instability arises due to gradient vanishing and mode collapse, inspiring variants such as Wasserstein GANs~\citep{arjovsky2017wasserstein}.

GANs can be incorporated to enhance other generative models. For example, Discriminator Guidance~\citep{kim2023refining} utilizes the gradient information from the discriminator as a corrective term to refine the score function in diffusion models (discussed in Section~\ref{sec:compare2}). Additionally, GANs are commonly employed as an auxiliary loss to improve one-step generation such as in diffusion distillation~\citep{kim2023consistency,yin2024improved,zhou2024adversarial}.
\section{Direct Discriminative Optimization}
\begin{figure*}[t]
    \centering
    \includegraphics[width=0.9\linewidth]{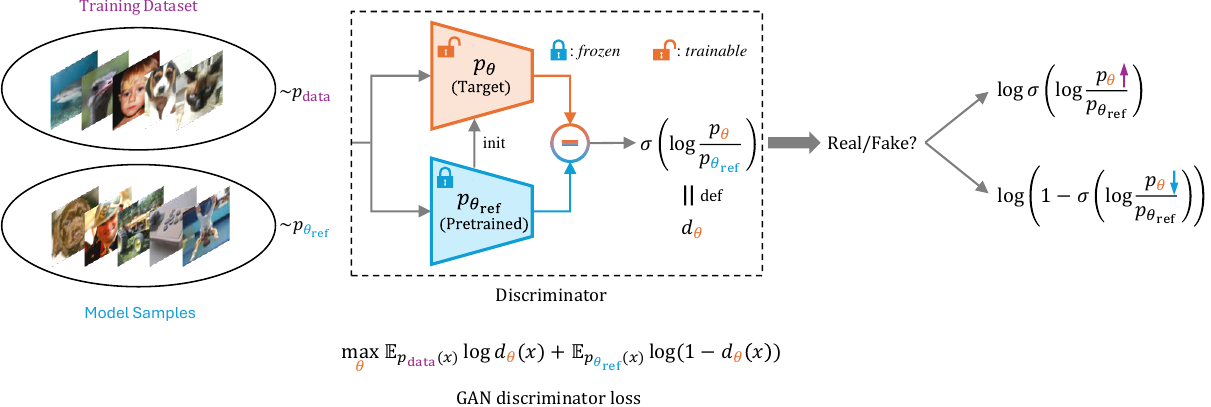}
    \vspace{-0.1in}
    \caption{\label{fig:pipeline} Illustration of DDO. (1) Models. $\tref$ is the (pretrained) reference model frozen during training. $\theta$ is the learnable model initialized as $\tref$.  (2) Data. Samples from $p_\data$ are drawn from the training dataset. Samples from $p_{\tref}$ are generated by the reference model, either offline or online. (3) Objective. The target model $\theta$ is optimized by applying the GAN discriminator loss with the implicitly parameterized discriminator $d_\theta$ to distinguish between real samples from $p_\data$ and fake samples from $p_{\tref}$. }
    \vspace{-0.1in}
\end{figure*}

Motivated by the benefits of adversarial training in enhancing generation quality, we aim to bridge likelihood-based generative models with GANs to derive an alternative training paradigm to MLE. Unlike prior works that incorporate GAN as an auxiliary loss and require additional engineering overhead, our approach seeks to (1) directly optimize likelihood-based generative models without modifying the network architecture, adding extra discriminators, complicating the training procedure or increasing inference costs, and (2) eliminate the need for backpropagation through the sampling process, making it applicable to diffusion and autoregressive models that rely on iterative sampling.

\subsection{Your Likelihood-Based Generative Model is Secretly a Discriminator}
Unlike one-step generators that learn a direct mapping from noise to data, likelihood-based generative models are grounded in the probabilistic definition of the likelihood function $p_\theta$, which enables both the generation of samples $\x\sim p_\theta$ and the evaluation of the likelihood $p_\theta(\x)$, either exactly or approximately, while retaining the tractability of backpropagation through the likelihood computation. This inspires us to utilize
the likelihood information embedded in the optimal discriminator (Eqn.~\eqref{eq:optimal-discriminator}).

Specifically, consider a pretrained model $p_{\tref}$ as a reference to generate fake samples. The optimal discriminator $d_\theta$ for
\begin{equation}
    \min_\theta-\E_{p_\data(\x)}\left[\log d_\theta(\x)\right]-\E_{p_{\tref}(\x)}\left[\log(1-d_\theta(\x))\right]
\end{equation}
can be rewritten as:
\begin{equation}
    d^*(\x)=\frac{p_\data(\x)}{p_\data(\x)+ p_{\tref}(\x)}=\sigma\left(\log\frac{p_\data(\x)}{p_{\tref}(\x)}\right)
\end{equation}
where $\sigma(x)=\frac{1}{1+e^{-x}}$ is the Sigmoid function. The data distribution $p_\data$ is available from $d^*$. Therefore, if we parameterize the discriminator $d_\theta$ using a likelihood-based target generative model $p_\theta$ as:
\begin{equation}
    d_\theta(\x)\coloneqq \sigma\left(\log\frac{p_\theta(\x)}{p_{\tref}(\x)}\right)
\end{equation}
then the optimal target model that minimizes the GAN discriminator loss matches the data distribution. We formalize this induced objective in the following theorem.
\begin{theorem}[Optimality]
\label{theorem:raw}
    With unlimited model capacity, the optimal likelihood-based model $p_\theta$ under the objective
    \begin{equation}
    \label{eq:ddo-loss1}
        \begin{aligned}
            \min_\theta \Lc(\theta)=&-\E_{p_\data(\x)}\left[\log \sigma\left(\log\frac{p_\theta(\x)}{p_{\tref}(\x)}\right)\right]\\
            &-\E_{p_{\tref}(\x)}\left[\log\left(1-\sigma\left(\log\frac{p_\theta(\x)}{p_{\tref}(\x)}\right)\right)\right]
        \end{aligned}
    \end{equation}
    satisfies $p_{\theta^*}=p_\data$.
\end{theorem}
In contrast to previous GAN-based methods that introduce a separate discriminator network $d_\phi$, our approach implicitly defines the discriminator through a target generative model $p_\theta$. While it is theoretically feasible to initialize $\theta,\tref$ arbitrarily and train from scratch, strong initial conditions facilitate optimization (Section~\ref{sec:theoretical-analysis}). In practice, we initialize $\theta,\tref$ from widely available pretrained models, promoting steady improvement. We refer to this approach as \textit{Direct Discriminative Optimization} (DDO), drawing parallels with Direct Preference Optimization (DPO)~\citep{rafailov2024direct}, which aligns language models with human preferences by expressing the reward model in terms of the likelihood ratio between two policies (discussed in Section~\ref{sec:compare1}). The DDO pipeline is illustrated in Figure~\ref{fig:pipeline}.

\textbf{What does the DDO update do?}\hspace{1em}For a mechanistic understanding of DDO, we can analyze the gradient of the loss function with respect to parameters $\theta$:
\begin{equation}
    \resizebox{0.48\textwidth}{!}{$\nabla_\theta\Lc(\theta)
    =\int (\underbrace{1-d_\theta(\x)}_{\in [0,1]})(\underbrace{p_\theta(\x)-p_\data(\x)}_{p_\theta(\x)\uparrow  \text{ when } <0})\nabla_\theta\log p_\theta(\x)\dm\x$}
\end{equation}
Intuitively, gradient descent increases the model likelihood $p_\theta(\x)$ for data points $\x$ that satisfy $p_\theta(\x)<p_\data(\x)$, and decreases it otherwise, pushing $p_\theta$ closer to $p_\data$. Furthermore, the gradient magnitude is weighted by both the distance $|p_\theta(\x)-p_\data(\x)|$ and $1-d_\theta(\x)$, assigning higher weights to samples discriminated as fake.
\subsection{Theoretical Analysis}
\label{sec:theoretical-analysis}
Apart from the optimality guarantee, we also examine the behavior of the DDO objective when $\theta$ is not optimal. Specifically, we investigate the following question:
\begin{align*}
\textit{Is $p_{\theta}$ closer to $p_\data$ with a lower $\Lc(\theta)$?}
\end{align*}
Under certain assumptions, we can establish bounds on the divergence between $p_\theta$ and $p_\data$ in terms of the difference between $\Lc(\theta)$ to the optimal loss value $\Lc^*$, as formalized in the following theorem.
\begin{theorem}[Divergence Bounds]
\label{theorem:bound}
    If $\log\frac{p_{\tref}}{p_\data}$ and $\log\frac{p_\theta}{p_{\tref}}$ are bounded, there exist some constants $C_1,C_2$ such that
    \begin{align}
        \kl{p_\data}{p_\theta}\leq C_1\sqrt{\Lc(\theta)-\Lc^*}\\
        \kl{p_\theta}{p_\data}\leq C_2\sqrt{\Lc(\theta)-\Lc^*}
    \end{align}
\end{theorem}
\vspace{-0.05in}
The assumption of bounded $\log\frac{p_\theta}{p_{\tref}}$ implies that the optimized distribution does not deviate significantly from the reference distribution, which is reasonable when finetuning for a short duration. The assumption of bounded $\log\frac{p_{\tref}}{p_\data}$ imposes a constraint to the reference model regarding its mutual density coverage with the data distribution. We can expect $\log\frac{p_{\tref}}{p_\data}$ to be lower bounded, i.e., $p_{\tref}$ sufficiently covers $p_\data$, which aligns with the characteristics of MLE-trained models. Under this condition, the forward KL $\kl{p_\data}{p_\theta}$ remains bounded by $\sqrt{\Lc(\theta)-\Lc^*}$. However, bounding the reverse KL requires an upper bound on $\frac{p_{\tref}}{p_\data}$, which imposes a stronger constraint on $p_{\tref}$.
\subsection{Practical Implementation}
We introduce several practical techniques that make DDO applicable to high-dimensional real-world data and diffusion models whose likelihood computation is expensive. \footnote{A code example is provided in Appendix~\ref{appendix:example}.}

\textbf{Generalized Objective with Extra Coeffecients}\hspace{1em}The log-likelihood $\log p_\theta(\x)$ of likelihood-based generative models often scales with the data dimension and can reach magnitudes of $10^3$. As the DDO objective in Eqn.~\eqref{eq:ddo-loss1} involves a Sigmoid operation on $\log p_\theta(\x)$, this can lead to gradient vanishing and numerical precision issues. To address this, we add hyperparameters $\alpha,\beta$ to control the relative weights of loss terms and scale the probability ratio:
\vspace{-0.05in}
\begin{equation}
\label{eq:ddo-loss2}
    \begin{aligned}
        \Lc_{\alpha,\beta}(\theta)=&-\E_{p_\data(\x)}\left[\log \sigma\left(\beta\log\frac{p_\theta(\x)}{p_{\tref}(\x)}\right)\right]\\
        &-\alpha\E_{p_{\tref}(\x)}\left[\log\left(1-\sigma\left(\beta\log\frac{p_\theta(\x)}{p_{\tref}(\x)}\right)\right)\right]
    \end{aligned}
\end{equation}
The modified loss retains the same optimization trend, namely, increasing $p_\theta(\x)$ for $\x\sim p_\data$ and decreasing $p_\theta(\x)$ for $\x\sim p_{\tref}$, but the optimum may ``overshoot'' the data distribution for $\beta<1$. Specifically, we have:
\begin{theorem}
\label{theorem:alpha-beta}
    With unlimited model capacity, the optimal likelihood-based generative model $\theta$ that minimizes $\Lc_{\alpha,\beta}(\theta)$ satisfies $p_{\theta^*}\propto p_{\tref}^{1-1/\beta}p_\data^{1/\beta}$ for certain $\alpha$.
\end{theorem}
This establishes a deep connection with guidance methods (discussed in Section~\ref{sec:compare2}). In practice, we observe that $\alpha$ and $\beta$ across a wide range of values\footnote{Typical choices are $\alpha\in[0.5,50]$ and $\beta\in[0.01,0.1]$, while the optimal values depend on the specific model and settings.} yield reasonable performance. We sweep over them for the best results.

\textbf{Handling Compute-Intensive Likelihood}\hspace{1em}Evaluating the model likelihood for a specific data point can be computationally intensive. In particular, unlike autoregressive models, which only require a single forward pass through the network to compute $\log p_\theta(\x)$ (Eqn.~\eqref{eq:ar_logp}) due to the causal structure imposed by attention masks, diffusion models necessitate multiple forward passes over different timesteps to approximate $\log p_\theta(\x)$ through the ELBO (Eqn.~\eqref{eq:diffusion_logp}). Specifically, the log-likelihood ratio in the DDO loss is:
\begin{equation}
    \log\frac{p_\theta(\x)}{p_{\tref}(\x)}\approx \E_{t,\epsilonv}\left[\Delta_{\x_t,t,\epsilonv}\right]
\end{equation}
where $\x_t=\alpha_t\x+\sigma_t\epsilonv$ and
\begin{equation}
    \Delta_{\x_t,t,\epsilonv}=- w(t)\left(\|\epsilonv_\theta(\x_t,t)-\epsilonv\|_2^2-\|\epsilonv_{\tref}(\x_t,t)-\epsilonv\|_2^2\right)
\end{equation}
We apply Jensen's inequality pointwise to derive an upper bound for the loss using the convexity of the function $-a\log\sigma(x)-b\log(1-\sigma(x))$ for any $a,b\geq 0$:
\begin{equation}
\label{eq:diffusion-ddo-approx}
\resizebox{0.48\textwidth}{!}{$
\begin{aligned}
    &\Lc(\theta)\\
    \approx&-\E_{p_\data(\x)}\log \sigma\left(\E_{t,\epsilonv}\left[\Delta\right]\right)-\E_{p_{\tref}(\x)}\log(1-\sigma\left(\E_{t,\epsilonv}\left[\Delta\right]\right))\\
    \leq &-\E_{t,\epsilonv}\left[\E_{p_\data(\x)}\log \sigma(\Delta)+\E_{p_{\tref}(\x)}\log(1-\sigma(\Delta)))\right]
\end{aligned}$
}
\end{equation}
This treatment, analogous to the one used in Diffusion-DPO~\citep{wallace2024diffusion}, enables us to approximate the diffusion DDO loss using a single forward pass for each $\x$.

\textbf{Multi-Round Refinement via Self-Play}\hspace{1em}Due to the practical modifications for applicability, the optimization process of DDO provides useful gradient information in the early stage but does not converge to the data distribution in the final. To maximize the fine-tuning performance, we adopt a multi-round refinement strategy, where the reference model $p_{\tref}$ is iteratively updated by replacing it with an improved version from the previous round:
\vspace{-0.05in}
\begin{smalleralign}
\nonumber
\text{Round $n$:}\quad \ldots &\quad  \rightarrow  \underbrace{p_{\theta^*_{n-1}}}_{\text{Reference}} \rightarrow  \quad \underbrace{\sigma\left(\beta \log \frac{p_{\theta_n}}{p_{\theta_{n-1}^*}}\right)}_{\text{Discriminator}} \\
\nonumber
\text{Round $n+1$:}\quad&\quad \rightarrow \underbrace{p_{\theta_n^*}}_{\text{Reference} } \rightarrow \quad \ldots
\end{smalleralign}
where $\theta_n^*$ represents the best-performing model across different hyperparameter configurations and training iterations in round $n$. In each round, the reference model acts as a fixed generator, making the multi-round optimization analogous to the generator-discriminator interplay in GANs. However, unlike GANs, where both networks are explicitly optimized, we never update the reference (generator) model directly. Instead, the generator is obtained from the discriminator in the previous round, leading to a form of self-play. This iterative refinement process is conceptually similar to Iterative DPO~\citep{xu2023some} and SPIN~\citep{chen2024self}, which extend DPO for better language model alignment. 

\subsection{Discussion}
\textbf{Connection to RL}\hspace{1em}At a high level, DDO enables visual generative models to \textbf{utilize negative signals from self-generated samples}—a characteristic deeply rooted in reinforcement learning (RL) that underpins modern language models~\citep{achiam2023gpt,guo2025deepseek}. Distinguished from works that employ a similar contrastive loss merely to off-the-shelf data~\citep{cca}, DDO can fundamentally improve the base model's ability.

\textbf{Extension to $f$-divergence}\hspace{1em}The GAN discriminator loss can be generalized to $f$-divergences~\citep{nowozin2016f}:
\begin{equation}
    D_f\infdivx{p}{q}=\sup_{T}\E_{p(x)}\left[T(x)\right]-\E_{q(x)}\left[f^*(T(x))\right]
\end{equation}
where $f^*$ is the convex conjugate of $f$. DDO can be extended to this family as the optimal $T^*(x)=f'\left(\frac{p(x)}{q(x)}\right)$ explicitly involves the density ratio.
\section{Comparison with Existing Methods}
\subsection{Direct Preference Optimization (DPO)}
\label{sec:compare1}
\begin{figure}[t]
    \centering
    \includegraphics[width=0.9\linewidth]{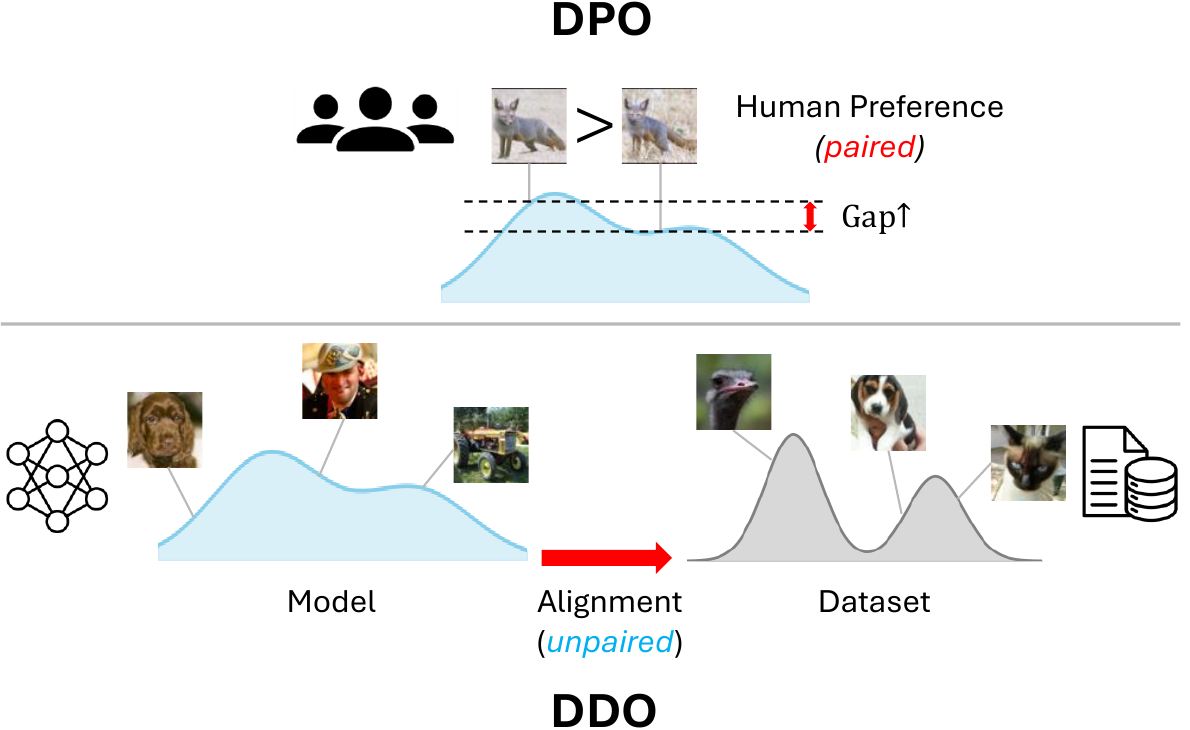}
    \vspace{-0.1in}
    \caption{\label{fig:dpo} Comparison with DPO.}
    \vspace{-0.1in}
\end{figure}
DPO~\citep{rafailov2024direct} is a lightweight surrogate objective designed for reinforcement learning from human feedback (RLHF)~\citep{ouyang2022training,achiam2023gpt} that enhances the instruction-following ability of pretrained language models. Standard RLHF involves two stages: (1) learning a reward model $r_\theta$ and (2) aligning the reference policy $\pi_{\tref}$ to the target policy $\pi_\theta(y|x)\propto \pi_{\tref}(y|x)e^{r_\theta(x,y)/\beta}$ using RL, where $x$ is the prompt and $y$ is the response. The Bradley-Terry preference mode~\citep{bradley1952rank} links preferences and rewards by
\begin{equation}
\resizebox{0.48\textwidth}{!}{$
    p(y_w\succ y_l|x)\coloneqq\frac{e^{r(x,y_w)}}{e^{r(x,y_l)}+e^{r(x,y_w)}}=\sigma(r(x,y_w)-r(x,y_l))$}
\end{equation}
where $y_w$ and $y_l$ denote the winning and losing responses for a given prompt $x$, annotated by human. DPO enables direct optimization of pretrained language models on preference data without training a separate reward model:
\begin{equation}
\resizebox{0.48\textwidth}{!}{$
\begin{aligned}
    &\Lc^{\rm DPO}(\theta)\\
    =&-\E_{(x,y_w,y_l)\sim\Dc}\log\sigma\left(\beta\log\frac{\pi_\theta(y_w|x)}{\pi_{\tref}(y_w|x)}-\beta\log\frac{\pi_\theta(y_l|x)}{\pi_{\tref}(y_l|x)}\right)
\end{aligned}$
}
\end{equation}
where the reward function $r_\theta(y|x)$ 
is implicitly parameterized by the log-likelihood ratio $\beta\log\frac{\pi_\theta(y|x)}{\pi_{\tref}(y|x)}$.

Despite sharing similar insights in parameterization, DDO is fundamentally different from DPO. As illustrated in Figure~\ref{fig:dpo}, DPO is designed for \textit{preference learning}, requiring additional paired human-annotated data and maximizing the likelihood gap between preferred (winning) and non-preferred (losing) responses without considering the whole distribution. In contrast, DDO focuses on \textit{distribution alignment}, directly aligning the model with the ground-truth data distribution. It requires only the original training data that are unpaired with the model-generated samples.
\subsection{Guidance Methods}
\label{sec:compare2}
\begin{figure}[t]
    \centering
    \includegraphics[width=1.0\linewidth]{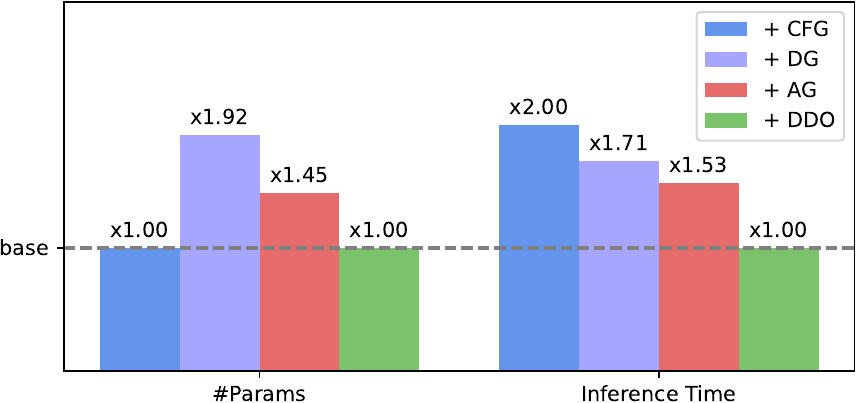}
    \vspace{-0.2in}
    \caption{\label{fig:compare} Comparison of model parameter counts and inference time across different guidance methods and DDO. For DG, we measure the statistics on class-conditional CIFAR-10. For AG, we measure the statistics on ImageNet-64.}
    \vspace{-0.2in}
\end{figure}
We review several types of guidance methods that enhance diffusion models\footnote{Guidance can be easily adapted to autoregressive models.} at inference time. Let $\sv_\theta(\x_t,t)$ denote the score function representation introduced in Section~\ref{sec:bg1}.

\textbf{Classifier-Free Guidance (CFG)}~\citep{ho2021classifier} combines the unconditional/conditional model to obtain a new score predictor $\sv_\theta'(\x_t,t,c)\coloneqq \sv_\theta(\x_t,t,c)+w(\sv_\theta(\x_t,t,c)-\sv_\theta(\x_t,t,\emptyset))$, where $\emptyset$ represents the unconditional case, and $w$ is the guidance scale. The unconditional model shares parameters with the conditional model and is learned by random label dropout during training.

\textbf{Discriminator Guidance (DG)}~\citep{kim2023refining} trains a time-dependent discriminator network $d_\phi$ that distinguishes between perturbed real data and model-generated samples. Its gradient is then used to refine the score function: $\sv_{\theta,\phi}(\x_t,t)=\sv_\theta(\x_t,t)+w\nabla_{\x_t}\log\frac{d_\phi(\x_t,t)}{1-d_\phi(\x_t,t)}$. Similar to DDO, DG 
leverages the optimal discriminator (Eqn.~\eqref{eq:optimal-discriminator}) for theoretical guarantees. However, it explicitly parameterizes the discriminator as a separate, time-aware network.

\textbf{Autoguidance (AG)}~\citep{karras2024guiding} operates similarly to CFG but refines the score function by guiding the base model with an inferior variant: $\sv_{\theta,\phi}(\x_t,t)\coloneqq \sv_\theta(\x_t,t)+w(\sv_\theta(\x_t,t)-\sv_\phi(\x_t,t))$, where $\sv_\phi$ is a degraded version of $\sv_\theta$ obtained via reduced model capacity or under-training.

For a unified perspective, all these guidance methods improve a well-trained distribution $p_\theta$ by amplifying its difference from a degraded or less informative distribution $p_\phi$: $p_{\theta,\phi}\propto p_\theta\left(\frac{p_\theta}{p_\phi}\right)^w$. This superposition sharpens the MLE-optimized model distribution and suppresses low-probability outliers~\citep{karras2024guiding}. According to Theorem~\ref{theorem:alpha-beta}, DDO with $\beta<1$ induces a similar overshooting effect, highlighting the theoretical connection. 

Unlike guidance methods, DDO enhances sample quality without increasing inference costs compared to the base model (Figure~\ref{fig:compare}). Moreover, in scenarios where CFG is crucial for balancing image-condition alignment and diversity (e.g., FID-IS curve), DDO can be seamlessly integrated with CFG to achieve an overall improved trade-off (Section~\ref{sec:exp2}).

\section{Experiments}
Our experiments aim to investigate the following aspects:
\vspace{-0.1in}
\begin{enumerate}
\item The effectiveness and efficiency of DDO in enhancing the visual quality of well-trained diffusion models (Section~\ref{sec:exp1}) and autoregressive (Section~\ref{sec:exp2}) models.
\vspace{-0.1in}
\item The impact of the hyperparameters $\alpha,\beta$, as well as the benefits of multi-round refinement.
\end{enumerate}
\vspace{-0.1in}
\subsection{Experimental Setups}
\textbf{Datasets \& Models}\hspace{1em}We experiment on standard image benchmarks including CIFAR-10~\citep{krizhevsky2009learning} in 32$\times$32 resolution and ImageNet~\citep{deng2009imagenet} in multiple resolutions (64$\times$64, 256$\times$256, 512$\times$512). For each dataset, we apply DDO to finetune state-of-the-art diffusion or autoregressive models, including EDM~\citep{karras2022elucidating}, EDM2~\citep{karras2024analyzing} and VAR~\citep{tian2024visual}. We compare with a range of advanced generative baselines, including diffusion models, autoregressive (AR) models, masked models, and GAN-based approaches.

\textbf{Training \& Evaluation}\hspace{1em}We evaluate Fréchet inception distance (FID)~\citep{heusel2017gans} on 50k images 
as the primary benchmark metric for all experiments, and additionally measure Inception Score (IS)~\citep{barratt2018note} for ImageNet 256$\times$256. We report the number of function evaluations (NFE) as a quantification of inference efficiency. We find strict class balance crucial for FIDs on ImageNet and slightly modify the original EDM sampling scripts to enforce it. For diffusion models, we finetune over multiple rounds until further improvement is negligible. For VAR, we observe rapid convergence and only finetune for 2 rounds. The finetuning is highly efficient, with each round requiring less than 1\% of pretraining iterations. Further experiment details can be found in Appendix~\ref{appendix:exp-details}, and visualizations of generated samples are provided in Appendix~\ref{appendix:additional-results}.
\subsection{Results on Diffusion Models}
\label{sec:exp1}
\begin{table}[!th]
\centering
\vspace{-0.1in}
\caption{\label{tab:cifar10}\small Results on unconditional and class-conditional CIFAR-10. $^\dagger$Including diffusion distillation methods with auxiliary GAN loss. $^\ddagger$ The reported parameter count excludes those of the discriminator (for GANs) and VAE encoder/decoder (for latent-space models).}
\vspace{2pt}
\resizebox{.48\textwidth}{!}{
\begin{tabular}{clcccc}
 \toprule
\multirow{2}{*}{Type} & \multirow{2}{*}{Model}&\multirow{2}{*}{\#Params$^\ddagger$}&\multirow{2}{*}{NFE} & \multicolumn{1}{c}{Uncond} & \multicolumn{1}{c}{Cond} \Bstrut\\
\cline{5-6}
&&& & FID$\downarrow$ &FID$\downarrow$ \Tstrut\\ 
 \midrule
 \multirow{5}{*}{GAN$^\dagger$} & StyleGAN2-ADA~\citep{karras2020training} &20M& 1 &2.92&2.42\\ %
 &StyleGAN-XL~\citep{sauer2022stylegan}&18M&1&-&1.85\\
 &R3GAN~\citep{huang2025gan}&21M&1&-&1.96\\
 &CTM~\citep{kim2023consistency}&59M&1&1.98&1.73\\
 &SiD$^2$A~\citep{zhou2024adversarial}&56M&1&1.50&1.40\\
 
 \midrule
 \multirow{12}{*}{Diffusion} & DDPM~\citep{ho2020denoising}&36M & 1000 &3.17&- \\
 &iDDPM~\citep{nichol2021improved} &50M& 4000&2.90&-\\
 &DDIM~\citep{ho2020denoising}&36M & 100 &4.16&- \\ 
 &DPM-Solver~\citep{lu2022dpm} &107M& 48 &2.65&- \\ 
 &DPM-Solver-v3~\citep{zheng2023dpm}&56M& 12 &2.24&- \\ 
 &NCSN++~\citep{song2020score}&108M&2000&2.20&-\\
 &LSGM~\citep{vahdat2021score}&376M&138&2.10&-\\
 &VDM~\citep{kingma2021variational}&31M&1000&7.41&-\\
 &Flow Matching~\citep{lipman2022flow}&-&142&6.35&-\\
 &i-DODE~\citep{zheng2023improved} &139M& 215 &3.76&-\\
 &EDM~\citep{karras2022elucidating} &56M& 35 &1.97&1.79 \\ 
 &+ DG~\citep{kim2023refining}&107M&53&1.77&1.64\\
 \midrule
 \multirow{2}{*}{Ours}&EDM (retested)&56M&35&1.97&1.85\\
 &+ DDO&56M&35&\textbf{1.38}&\textbf{1.30}\\
 \bottomrule
\end{tabular}
}
\vspace{-4.5mm}
\end{table}

\begin{table}[!th]
\centering

\caption{\label{tab:imagenet64}\small 
Results on class-conditional ImageNet-64.}
\vspace{2pt}
\resizebox{.425\textwidth}{!}{
\begin{tabular}{clccc}
 \toprule
Type & Model&\#Params&NFE & FID$\downarrow$ \\ 
 \midrule
 \multirow{6}{*}{GAN} & StyleGAN-XL~\citep{sauer2022stylegan} &135M& 1 &1.51\\

 &R3GAN~\citep{huang2025gan}&104M&1&2.09\\
 &CTM~\citep{kim2023consistency}&324M&1&1.92\\
 &DMD2~\citep{yin2024improved}&296M&1&1.28\\
 &PaGoDA~\citep{kim2024pagoda}&296M&1&1.21\\
 &SiD$^2$A~\citep{zhou2024adversarial}&296M&1&1.11\\
 
 \midrule
 \multirow{12}{*}{Diffusion} 
 &iDDPM~\citep{nichol2021improved} &270M& 250&2.92\\
 &ADM~\citep{dhariwal2021diffusion}&296M&250&2.07\\
 &RIN~\citep{jabri2022scalable}&281M&1000&1.23\\
 &EDM~\citep{karras2022elucidating}&296M&511&1.36\\
 &VDM++~\citep{kingma2024understanding}&296M&511&1.43\\
 &DisCo-Diff~\citep{xu2024disco}&-&623&1.22\\
 &EDM2-S~\citep{karras2024analyzing}&280M&63&1.58\\
 &+ CFG~\citep{ho2021classifier}&560M&126&1.48\\
 &+ AG~\citep{karras2024guiding}&405M&126&1.01\\
 &EDM2-M&498M&63&1.43\\
 &EDM2-L&777M&63&1.33\\
 &EDM2-XL&1.1B&63&1.33\\
 \midrule
 \multirow{2}{*}{Ours}&EDM2-S (retested)&280M&63&1.60\\
 &+ DDO&280M&63&\textbf{0.97}\\
 \bottomrule
\end{tabular}
}
\vspace{-3mm}
\end{table}

\begin{table}[!th]
\centering
\vspace{-0.1in}
\caption{\label{tab:imagenet512}\small Results on class-conditional ImageNet 512$\times$ 512. ``G'' denotes guidance (CFG by default). The reported NFE values are without guidance; applying guidance doubles them. $^\dagger$With classifier guidance. $^\ddagger$With autoguidance (AG)~\citep{karras2024guiding}.}
\vspace{2pt}
\resizebox{.48\textwidth}{!}{
\begin{tabular}{clcccc}
 \toprule
\multirow{2}{*}{Type} & \multirow{2}{*}{Model}& \multicolumn{3}{c}{w/o G} & \multicolumn{1}{c}{w/ G} \Bstrut\\
\cline{3-6}
&&\#Params&NFE & FID$\downarrow$ &FID$\downarrow$ \Tstrut\\
 \midrule
 \multirow{3}{*}{GAN} &BigGAN-deep~\citep{brock2018large} &112M&1&8.43&-\\
 &StyleGAN-XL~\citep{sauer2022stylegan} &168M& 1 &2.41&-\\ %
 &SiD$^2$A~\citep{zhou2024adversarial}& 1.5B&1&1.37&-\\
 
 \midrule
 \multirow{15}{*}{Diffusion} & ADM~\citep{dhariwal2021diffusion} &559M& 250 & 23.24&7.72$^\dagger$ \\
 &ADM-U &731M& 500 & 9.96& 3.85$^\dagger$ \\
 & DiT-XL/2~\citep{peebles2023scalable} &675M& 250 & 12.03 & 3.04 \\
 & SiT-XL~\citep{ma2024sit} &675M&250  &-&2.62 \\
&+ REPA~\citep{yu2024representation}&675M&250&-&2.08\\
&RIN~\citep{jabri2022scalable}&410M&1000&3.95&-\\
 &U-ViT, L~\citep{hoogeboom2023simple}&2B&512&3.54&3.02\\
 &VDM++~\citep{kingma2024understanding}&2B&512&2.99&2.65\\
 &USiT-2B~\citep{chen2024deep}&2B&-&2.90&1.72\\
 &EDM2-XS~\citep{karras2024analyzing}&125M&63&3.53&2.91\\
 &EDM2-S&280M&63&2.56&2.23\\
 &+ AG&&&&1.34$^\ddagger$\\
 &EDM2-L&777M&63&2.06&1.88\\
 &EDM2-XXL&1.5B&63&1.91&1.81\\
 &+ AG&&&&1.25$^\ddagger$\\
 \midrule
 \multirow{3}{*}{Masked} & MaskGIT~\citep{chang2022maskgit} & 227M&12 &7.32 & -\\
 &MAGVIT-v2~\citep{yu2023language} &307M&64&3.07& 1.91 \\
 &MAR-L~\citep{li2024autoregressive} &481M&1024 &2.74 &1.73 \\
 \midrule
 \multirow{1}{*}{AR} 
 & VAR-d36-s~\citep{tian2024visual} & 2.3B&10&-&2.63 \\
 \midrule
 \multirow{3}{*}{Ours}
 &EDM2-L (retested)&777M&63&1.96&1.77\\
 &+ DDO&777M&63&\textbf{1.26}&\textbf{1.21}$^\ddagger$\\
 &\quad+ DPM-Solver-v3~\citep{zheng2023dpm}&777M&25&1.29&\textbf{1.21}$^\ddagger$\\
 \bottomrule
\end{tabular}
}
\vspace{-0.2in}
\end{table}

\begin{figure*}[t]

	\centering
	\begin{minipage}{.33\linewidth}
		\centering
			\includegraphics[width=\linewidth]{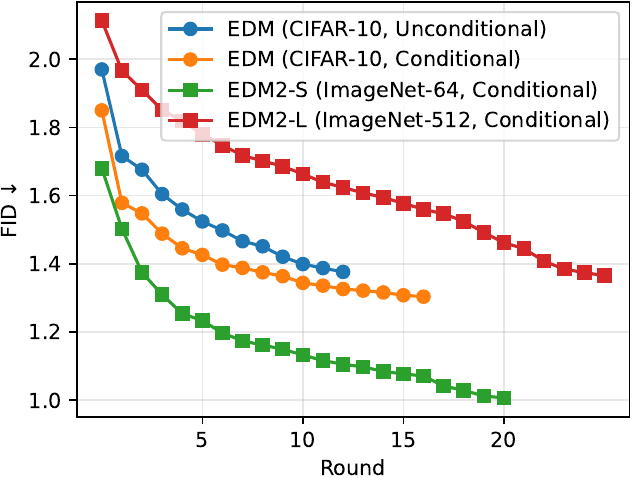}\\
\small{(a) FID-Round}
	\end{minipage}
	\begin{minipage}{.33\linewidth}
	\centering
	\includegraphics[width=\linewidth]{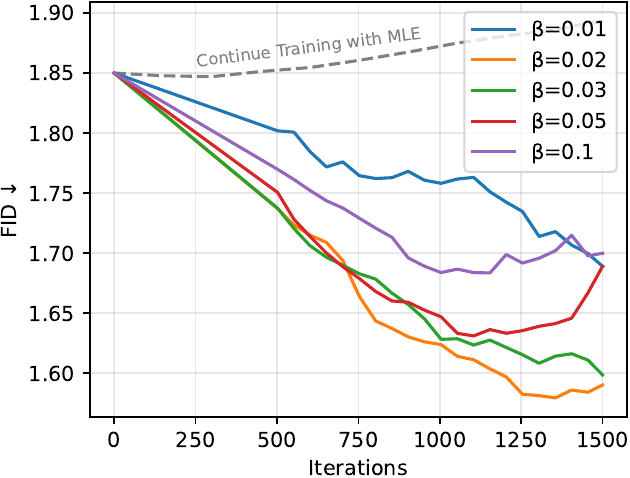}\\
\small{(b) $\alpha=4.0$}
\end{minipage}
	\begin{minipage}{.33\linewidth}
	\centering
	\includegraphics[width=\linewidth]{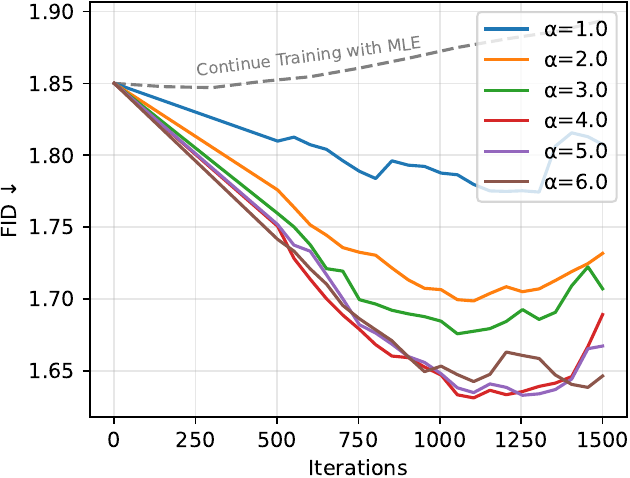}\\
\small{(c) $\beta=0.05$}
\end{minipage}
   \vspace{-.15in}
	\caption{\label{fig:diffusion} Illustrations of DDO on diffusion models, including (a) multi-round refinement and (b)(c) 
training curves under different $\alpha,\beta$. The ImageNet FIDs in this figure are evaluated without class rebalancing.}
	\vspace{-.15in}
\end{figure*}
The EDM and EDM2 base models on CIFAR-10 and ImageNet are implemented as separate unconditional or class-conditional networks. Since CFG provides limited benefits on these datasets, we directly apply the diffusion DDO loss without considering the interaction with CFG.

\textbf{Main Results}\hspace{1em}Table~\ref{tab:cifar10}, Table~\ref{tab:imagenet64} and Table~\ref{tab:imagenet512} present the quantitative results on CIFAR-10 and ImageNet. Figure~\ref{fig:diffusion}(a) illustrate the FID reduction over multiple rounds. We highlight the advantages of DDO as follows:

\textit{(1) Effectiveness.}  With multi-round refinement, DDO achieves record-breaking guidance-free FID scores of 1.38/1.30 on CIFAR-10, 0.97 on ImageNet-64, and 1.26 on ImageNet 512$\times$512, significantly improving upon the EDM and EDM2 base models by 30\%, 40\% and 36\%, respectively. Additionally, DDO outperforms all guidance-based and GAN-based methods requiring complex GAN-specific tuning or increasing inference costs.

\textit{(2) Efficiency.} Although we employ substantial training over dozens of rounds 
to maximize the performance and explore the upper bound of DDO, FID improves significantly within just a few rounds. Notably, DDO in a single round achieves FIDs of 1.72/1.58 on CIFAR-10 with the EDM base model, surpassing DG. On ImageNet-64, the compact EDM2-S attains an FID of 1.31 after only 3 rounds, surpassing the 4$\times$ larger EDM2-XL. On ImageNet 512$\times$512, guidance-free EDM2-L matches CFG-enhanced EDM2-XXL (2$\times$ larger) with only 4 rounds, demonstrating the parameter efficiency unlocked by DDO.

\textbf{Effects of $\alpha,\beta$}\hspace{1em}As shown in Figure~\ref{fig:diffusion}(b)(c), we visualize the training curves under different $\alpha,\beta$ for class-conditional CIFAR-10 during the first round. We empirically find that a wide range of $\alpha,\beta$ consistently improves the base model, though identifying the optimal hyperparameters requires grid searching. Moreover, tuning $\alpha$ while keeping $\beta$ fixed or adjusting $\beta$ under appropriate $\alpha$ yields similar effects.

\textbf{Comparison to MLE}\hspace{1em}Figure~\ref{fig:diffusion}(b)(c) also includes the FID curve of extended training using the original diffusion loss (i.e., MLE). Unlike DDO, continued MLE training fails to improve performance and even leads to degradation when we do not retain previous optimizer states. This is partly because the base model is already extremely optimized and more fundamentally due to the forward KL objective's inherent limitations. In contrast, DDO with various $\alpha, \beta$ configurations consistently demonstrates clear advancements.

\textbf{Accelerate Sampling}\hspace{1em} By leveraging the advanced sampler DPM-Solver-v3~\citep{zheng2023dpm}, we reduce the inference steps of EDM2-L+DDO to 25 while preserving generation quality. Our model achieves FID scores of 1.29/1.21 for 512×512 images, with an end-to-end per-image latency of 1.03s/1.96s on a single H100 GPU (batch size = 1).
\subsection{Results on Autoregressive Models}
\label{sec:exp2}
\begin{table}[!th]
\centering
\vspace{-0.1in}
\caption{\label{tab:imagenet256}\small Results on class-conditional ImageNet 256$\times$ 256. $^*$Samples are generated at $512\times512$ resolution and resized to $256\times256$ using OpenCV's \texttt{INTER\_LANCZOS4} method.}
\vspace{2pt}
\resizebox{.48\textwidth}{!}{
\begin{tabular}{clcccc}
 \toprule
\multirow{2}{*}{Type} & \multirow{2}{*}{Model}& \multicolumn{3}{c}{w/o G} & \multicolumn{1}{c}{w/ G} \Bstrut\\
\cline{3-6}
&&\#Params&NFE & FID$\downarrow$ &FID$\downarrow$ \Tstrut\\
 \midrule
 \multirow{3}{*}{GAN} &BigGAN-deep~\citep{brock2018large} &112M&1&6.95&-\\
 &GigaGAN~\citep{kang2023scaling}& 569M&1&3.45&-\\
 &StyleGAN-XL~\citep{sauer2022stylegan} &166M& 1 &2.30&-\\ %
 
 \midrule
 \multirow{10}{*}{Diffusion} & ADM~\citep{dhariwal2021diffusion} &554M& 250 & 10.94&4.59 \\
 &ADM-U &608M& 500 & 7.49&3.94 \\
 & LDM-4~\citep{rombach2022high} &400M& 250 & 10.56&3.60 \\
 & DiT-XL/2~\citep{peebles2023scalable} &675M& 250 & 9.62 & 2.27 \\
 & SiT-XL~\citep{ma2024sit} &675M&250  &8.3 &2.06 \\
&+ REPA~\citep{yu2024representation}&675M&250&5.90&1.42\\
&RIN~\citep{jabri2022scalable}&410M&1000&3.42&-\\
 &U-ViT, L~\citep{hoogeboom2023simple}&2B&512&2.77&2.44\\
 &VDM++~\citep{kingma2024understanding}&2B&512&2.40&2.12\\
 &LightningDiT~\citep{yao2025reconstruction}&675M&250&2.17&1.35\\
 \midrule
 \multirow{4}{*}{Masked} & MaskGIT~\citep{chang2022maskgit} & 227M&8 &6.18 & -\\
 &MAGVIT-v2~\citep{yu2023language} &307M&64&3.65& 1.78 \\
 &MAR-H~\citep{li2024autoregressive} &943M&256 &2.35 &1.55 \\
 &MaskBit~\citep{weber2024maskbit}&305M&256&-&1.52\\
 \midrule
 \multirow{9}{*}{AR} & VQGAN~\citep{esser2021taming} &1.4B& 256 & 15.78&- \\
 & ViT-VQGAN~\citep{yu2021vector} &1.7B&1024 &4.17 &- \\
 & RQ-Transformer~\citep{lee2022autoregressive} & 3.8B&68 &7.55 &- \\
 & LlamaGen-3B~\citep{sun2024autoregressive} &3.1B&256 & 13.58 &3.05 \\
 & Open-MAGVIT2-XL~\citep{luo2024open} &1.5B&256 &9.63 &2.33 \\
 & VAR-d16~\citep{tian2024visual} &310M& 10 &3.62 &3.30 \\
 & VAR-d30 &2.0B& 10 &2.17 & 1.90 \\
 &RAR-XXL~\citep{yu2024randomized}& 1.5B&256&3.91&1.48\\
 &xAR-H~\citep{ren2025beyond}&1.1B&50&-&1.24\\
 \midrule
 \multirow{7}{*}{Ours}
 &VAR-d16 (w/o sampling tricks)&310M&10&11.33&3.71\\
 &+ DDO&310M&10&\textbf{3.12}&\textbf{2.54}\\
 &VAR-d30 (w/o sampling tricks)&2.0B&10&4.74&1.92\\
 &+ CCA~\citep{cca}&2.0B&10&2.54&-\\
 &+ DDO&2.0B&10&\textbf{1.79}&\textbf{1.73}\Bstrut\\
 \cdashline{2-6}
 &EDM2-L + DDO (\textit{downsampled}$^*$)&777M&63&\textbf{1.26}&\textbf{1.21}\Tstrut\\
 &+ DPM-Solver-v3~\citep{zheng2023dpm}&777M&25&1.29&\textbf{1.21}\\
 \bottomrule
\end{tabular}
}
\vspace{-3mm}
\end{table}

\begin{figure*}[t]

	\centering
	\begin{minipage}{.25\linewidth}
		\centering
			\includegraphics[width=\linewidth]{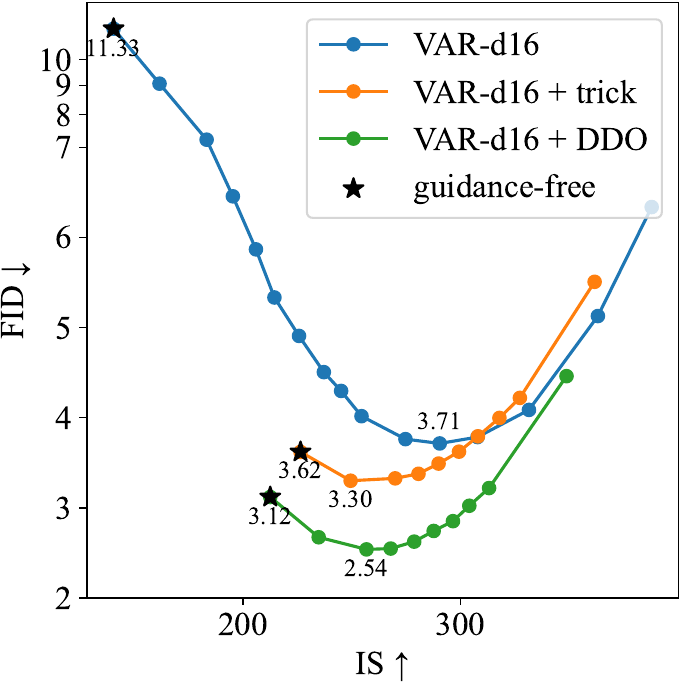}\\
\small{(a) VAR-d16}
	\end{minipage}
	\begin{minipage}{.25\linewidth}
	\centering
	\includegraphics[width=\linewidth]{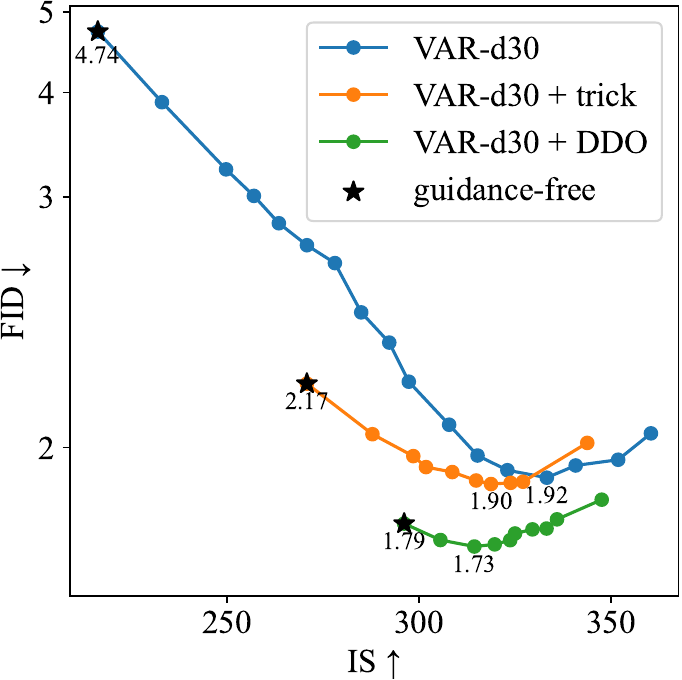}\\
\small{(b) VAR-d30}
\end{minipage}
	\begin{minipage}{.48\linewidth}
	\centering
	\includegraphics[width=0.92\linewidth]{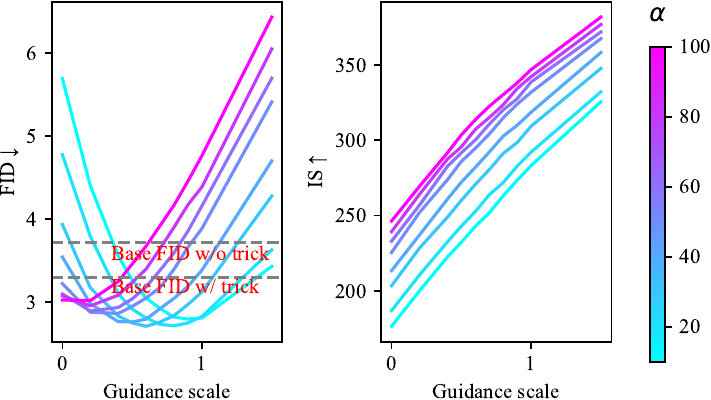}\\
\small{(c) FID and IS w.r.t. guidance scale $w$}
\end{minipage}
   \vspace{-.05in}
	\caption{\label{fig:ar} Illustrations of DDO on autoregressive models. (a)(b) FID-IS trade-off curves and (c) the impact of $\alpha$ under $\beta=0.02$.}
	\vspace{-.15in}
\end{figure*}
The VAR models rely heavily on CFG to enhance generation quality. A distinctive feature of CFG is its ability to balance diversity and fidelity by adjusting the guidance scale, which is essential for creating the FID-IS trade-off. Consequently, we need to accommodate DDO to ensure that the finetuned models remain compatible with CFG. To this end, we choose the reference and target distributions $p_{\tref},p_\theta$ as the guidance-free model corresponding to $w=0$. To preserve the model’s ability for unconditional generation, we incorporate random label dropout during DDO fine-tuning. We set $\alpha=0$ for the unconditional part to prevent it from receiving negative signals from reference samples $\x\sim p_{\tref}$.

\textbf{Main Results}\hspace{1em}Table~\ref{tab:imagenet256} presents the quantitative results on ImageNet 256$\times$256. Figure~\ref{fig:ar}(a)(b) illustrates the FID-IS trade-off varying the CFG scale. We summarize the advantages of DDO as follows:

\textit{(1) Eliminating sampling tricks.} It is worth noting that the original VAR results are based on top-$k$ and top-$p$ sampling strategies, which artificially lower the temperature. These heuristics introduce a training-inference gap and fail to reflect the genuine capability of pretrained models. In contrast, when evaluating models finetuned with DDO, we discard all such tricks, ensuring a more principled assessment of generative performance.

\textit{(2) Guidance-free performance.} DDO significantly reduces the guidance-free FID from 11.33/4.74 to 3.12/1.79 for VAR-d16 and VAR-d30, achieving 3.6$\times$ and 2.6$\times$ improvement. Notably, the guidance-free FIDs even outperform CFG-enhanced FIDs (3.30/1.90) of the original VAR results, indicating that higher-quality samples can be generated at half the inference cost with DDO. This is also superior to methods like CCA~\citep{cca} which only aim to remove the CFG but harm the model performance.

\textit{(3) CFG-enhanced performance.} When combined with CFG, the finetuned VAR models achieve significantly better FID-IS trade-offs than the pretrained counterparts, even when the latter employ sampling tricks. The lowest FID improves from 3.30/1.90 to 2.54/1.79. Furthermore, the finetuned VAR-d16 outperforms the 2$\times$ larger VAR-d20 (600M parameters) which has a CFG-enhanced FID of 2.57.

\textbf{Effects of $\alpha,\beta$}\hspace{1em}Figure~\ref{fig:ar}(c) visualizes FID and IS varying the CFG scale, where we finetune VAR-d16 under $\beta=0.02$ and different $\alpha$ for 60 iterations. The results indicate that all $\alpha \in [10,100]$ consistently achieve a CFG-enhanced FID lower than that of the base model. Larger values of $\alpha$ tend to yield lower guidance-free FIDs but may slightly weaken performance when combined with CFG.
\section{Conclusion}
In this work, we introduce Direct Discriminative Optimization (DDO), a universal enhancement technique designed for visual likelihood-based generative models. Inspired by the GAN framework and the parameterization insights from DPO, DDO breaks the curse of forward KL and substantially improves generation quality. Extensive experiments demonstrate its remarkable effectiveness and efficiency, surpassing state-of-the-art diffusion and autoregressive models and achieving record-breaking FID scores on standard image benchmarks. There remain promising directions for future exploration, such as eliminating the need for hyperparameter searching, improving inference efficiency through distillation, and scaling to tasks like text-to-image generation. We leave such avenues for future work.




\section*{Impact Statement}
This paper presents work whose goal is to advance the field of 
Machine Learning. There are many potential societal consequences
of our work, none of which we feel must be specifically highlighted here.


\bibliography{example_paper}
\bibliographystyle{icml2025}

\newpage
\appendix
\onecolumn
\section{Related Work}
\paragraph{Paradigms of Generative Models} Beyond autoregressive (AR) and diffusion models introduced in Section~\ref{sec:bg1}, other likelihood-based generative models have been explored in the literature. Variational autoencoders (VAEs)~\citep{kingma2013auto}, energy-based models~\citep{du2019implicit}, and normalizing flows~\citep{dinh2016density} are once popular for generative modeling of continuous data, but have since fallen out of favor due to their limited expressiveness and lack of scalability in modern large-scale generation tasks. In particular, VAEs are now primarily used as dimensionality reduction tools that compress data into latent spaces~\citep{esser2021taming,rombach2022high}, rather than generating samples from scratch. For discrete data generation, masked models, such as BERT~\citep{devlin2019bert}, CMLM~\citep{ghazvininejad2019mask} for masked language modeling and MaskGIT~\citep{chang2022maskgit} for masked image generation, offer an alternative likelihood-based paradigm to AR, characterized by bidirectional attention layers and random-order generation in place of AR's causal attention and raster order. Although the masked case of discrete diffusion models~\citep{austin2021structured,lou2023discrete,shi2024simplified,sahoo2024simple} has recently regained interest, it actually has little to do with continuous-space diffusion models and instead proves equivalent to the simpler and long-established masked models, while additionally suffering from hidden numerical precision issues that lower the effective temperature and lead to unfair evaluation when measured by the generative perplexity metric~\citep{zheng2024masked}, suggesting that the notion of ``diffusion'' is unnecessary in masked diffusion\footnote{Apart from the masked case, other discrete diffusion variants, such as uniform, have a more genuine connection to diffusion~\citep{sahoo2025diffusion}.}, serving as a somewhat hyped-up concept. There have also been pioneering efforts to combine different generative paradigms. MAR~\citep{li2024autoregressive} integrates masked modeling with the diffusion loss, enabling autoregressive image generation with continuous tokens. Transfusion~\citep{zhou2024transfusion} and Show-o~\citep{xie2024show} combine AR with diffusion/masked models for multi-modal generation, effectively synthesizing a mixture of text and image data. DDO is potentially applicable to these models for quality enhancement, and we leave such explorations for future work. 
\paragraph{Improving Generation Quality with GAN} Except for using GAN as an auxiliary loss for enhancing one-step or few-step generation in diffusion distillation~\citep{kim2023consistency,yin2024improved,zhou2024adversarial}, as mentioned in Section~\ref{sec:bg2}, several works have explored directly integrating diffusion models with GANs. Notably, \citet{xiao2021tackling} replaces the reverse denoising steps in diffusion models with a sequence of conditional GAN generators, enabling few-step generation. \citet{wang2022diffusion} modifies the GAN discriminator to distinguish between diffused real and generated samples in a time-aware manner. \citet{kim2023refining} leverages the gradient information from a trained discriminator to refine pretrained diffusion models. \citet{cca} adopts a binary classification loss with likelihood ratio parameterization similar to our objective, but its applicability is limited to removing CFG in autoregressive models and degrades the model performance. Apart from quality, quantized or sparse attention~\citep{zhang2025sageattention,zhang2024sageattention2,zhang2025sageattention3,zhang2025spargeattn} can be employed in parallel to accelerate inference.
\section{Theoretical Analyses of the DDO Objective}
In this section, we investigate the theoretical properties of the DDO objective and provide informal proofs to Theorem~\ref{theorem:raw}, Theorem~\ref{theorem:bound}, and Theorem~\ref{theorem:alpha-beta} in the main text.
\subsection{Analyses of $\Lc(\theta)$}
\paragraph{Optimal Solution} It is straightforward to show that the optimal $\theta$ minimizing $\Lc(\theta)$ satisfies $p_{\theta^*}=p_\data$ following the common GAN literature~\citep{goodfellow2014generative}. Specifically, let $r_\theta(\x)\coloneqq\log\frac{p_\theta(\x)}{p_{\tref}(\x)}$ denote the log-likelihood ratio between the learnable and reference distribution. The objective $\Lc(\theta)$ can be expressed as an integral form:
\begin{equation}
    \Lc(\theta)=\int \Lc(\theta)_{\x}\dm\x
\end{equation}
where 
\begin{equation}
    \Lc(\theta)_{\x}=-p_\data(\x)\log\sigma(r_\theta(\x))-p_{\tref}(\x)\log(1-\sigma(r_\theta(\x)))>0
\end{equation}
is the pointwise loss, and we only consider $\x$ in the valid range where $p_\data$ and $p_{\tref}$ have nonzero support. For any $(a,b)\in\R^2\backslash\{(0,0)\}$, the function $y\rightarrow -a\log y-b\log (1-y), y\in [0,1]$ achieves its minimum at $\frac{a}{a+b}$. Applying this to the pointwise loss $\Lc(\theta)_{\x}$, the minimizer satisfies
\begin{equation}
    \sigma(r_{\theta^*}(\x))=\frac{p_{\theta^*}(\x)}{p_{\theta^*}(\x)+p_{\tref}(\x)}=\frac{p_\data(\x)}{p_\data(\x)+p_{\tref}(\x)}\Rightarrow p_{\theta^*}(\x)=p_\data(\x)
\end{equation}
Since the global minimizer of $\Lc(\theta)$ is the pointwise minimizer of $\Lc(\theta)_{\x}$ for all $\x$, it follows that $p_{\theta^*}=p_\data$.
\paragraph{Loss Gradient} Using the derivative identity $\frac{\dm\sigma(x)}{\dm x}=\sigma(x)(1-\sigma(x))$, we obtain $\frac{\dm\log \sigma(x)}{\dm x}=1-\sigma(x), \frac{\dm\log(1-\sigma(x))}{\dm x}=-\sigma(x)$. Applying these to the pointwise loss, we derive the gradient w.r.t. $r_\theta(\x)$:
\begin{equation}
    \frac{\dm\Lc(\theta)_{\x}}{\dm r_\theta(\x)}=p_{\tref}(\x)\sigma(r_\theta(\x))-p_\data(\x)(1-\sigma(r_\theta(\x)))=\frac{p_\theta(\x)-p_\data(\x)}{p_\theta(\x)+p_{\tref}(\x)}p_{\tref}(\x)=(1-d_\theta(\x))(p_\theta(\x)-p_\data(\x))
\end{equation}
Thus, the full loss gradient is given by
\begin{equation}
        \nabla_\theta\Lc(\theta)=\int\nabla_\theta\Lc(\theta)_{\x}\dm\x
        =\int\frac{\dm\Lc(\theta)_{\x}}{\dm r_\theta(\x)}\nabla_\theta r_\theta(\x)\dm\x
        =\int(1-d_\theta(\x))(p_\theta(\x)-p_\data(\x))\nabla_\theta \log p_\theta(\x)\dm\x
\end{equation}
\paragraph{Divergence Bounds} We aim to derive bounds for the divergence between $p_\theta$ and $p_\data$ when $\theta$ is not optimal, using the difference between the loss value $\Lc(\theta)$ and its optimal counterpart $\Lc^*$. Without any assumptions, the forward KL divergence $\kl{p_\data}{p_\theta}$ is lower bounded by $\Lc(\theta)-\Lc^*$. By definition, $\Lc^*$ is the minimum loss value achieved when $p_\theta=p_\data$. The difference $\Lc(\theta)-\Lc^*$ can be decomposed as follows:
\begin{equation}
\begin{aligned}
    \Lc(\theta)-\Lc^*=&-\int p_\data(\x)\log\frac{p_\theta(\x)}{p_\theta(\x)+p_{\tref}(\x)}+p_{\tref}(\x)\log\frac{p_{\tref}(\x)}{p_\theta(\x)+p_{\tref}(\x)}\dm\x\\
    &+\int p_\data(\x)\log\frac{p_\data(\x)}{p_\data(\x)+p_{\tref}(\x)}+p_{\tref}(\x)\log\frac{p_{\tref}(\x)}{p_\data(\x)+p_{\tref}(\x)}\dm\x\\
    =&\int p_\data(\x)\log\frac{p_\data(\x)}{p_\theta(\x)}+(p_\data(\x)+p_{\tref}(\x))\log\frac{p_\theta(\x)+p_{\tref}(\x)}{p_\data(\x)+p_{\tref}(\x)} \dm\x\\
    =&\kl{p_\data}{p_\theta}-\kl{\frac{p_\data+p_{\tref}}{2}}{\frac{p_\theta+p_{\tref}}{2}}
\end{aligned}
\end{equation}
Therefore, $\kl{p_\data}{p_\theta}=\Lc(\theta)-\Lc^*+\kl{\frac{p_\data+p_{\tref}}{2}}{\frac{p_\theta+p_{\tref}}{2}}\geq \Lc(\theta)-\Lc^*$. While this result establishes a lower bound for the divergence, additional assumptions are required to derive an upper bound. Specifically, we assume that $\log\frac{p_{\tref}}{p_\data}$ and $\log\frac{p_\theta}{p_{\tref}}$ are bounded, i.e., there exist constants $M,M_1,M_2$ such that $\left|r_\theta(\x)\right|=\left|\log\frac{p_\theta(\x)}{p_{\tref}(\x)}\right|\leq M,M_1\leq\log\frac{p_{\tref}(\x)}{p_\data(\x)}\leq M_2$ for all $\x$. The pointwise loss can be expressed as a function of $r_\theta(\x)$:
\begin{equation}
    \Lc(\theta)_{\x}=f(r_\theta(\x))
\end{equation}
where
\begin{equation}
    f(y)\coloneqq -p_\data(\x)\log\sigma(y)-p_{\tref}(\x)\log(1-\sigma(y))=(p_\data(\x)+p_{\tref}(\x))\log\left(1+e^y\right)-p_\data y
\end{equation}
The first and second order derivatives of $f$ are given by:
\begin{equation}
    f'(y)=\frac{p_{\tref}(\x)e^{y}-p_\data(\x)}{1+e^{y}},\quad f''(y)=\frac{(p_\data(\x)+p_{\tref}(\x))e^{y}}{(1+e^{y})^2}=\frac{p_\data(\x)+p_{\tref}(\x)}{2+e^y+e^{-y}}
\end{equation}
Applying Taylor's expansion at $y=r_{\theta^*}(\x)=\log\frac{p_\data(\x)}{p_{\tref}(\x)}$, we obtain:
\begin{equation}
    f(r_{\theta}(\x))=f(r_{\theta^*}(\x))+f'(r_{\theta^*}(\x))(r_{\theta}(\x)-r_{\theta^*}(\x))+\frac{1}{2}f''(\xi)(r_{\theta}(\x)-r_{\theta^*}(\x))^2
\end{equation}
where $\xi\in[\min\{r_\theta(\x),r_{\theta^*}(\x)\},\max\{r_\theta(\x),r_{\theta^*}(\x)\}]$. Since $f'(r_{\theta^*}(\x))=0$ and $r_{\theta}(\x)-r_{\theta^*}(\x)=\log\frac{p_\theta(\x)}{p_\data(\x)}$, we get:
\begin{equation}
    \left(\log\frac{p_\theta(\x)}{p_\data(\x)}\right)^2=\frac{2}{f''(\xi)}(\Lc(\theta)_{\x}-\Lc(\theta^*)_{\x})
\end{equation}
Note that $f''(y)$ is a monotonically decreasing function w.r.t. $|y|$ and attains its maximum at the boundary of the given range, we have:
\begin{equation}
\begin{aligned}
    \frac{2}{f''(\xi)}&\leq\max\left\{\frac{2}{f''(r_{\theta}(\x))},\frac{2}{f''(r_{\theta^*}(\x))}\right\}\leq \max\left\{\frac{2}{f''(M)},\frac{2}{f''(r_{\theta^*}(\x))}\right\}\\
    &=2\max\left\{\frac{2+e^{-M}+e^M}{p_\data(\x)+p_{\tref}(\x)},\frac{p_\data(\x)+p_{\tref}(\x)}{p_\data(\x)p_{\tref}(\x)}\right\}
\end{aligned}
\end{equation}
Therefore,
\begin{equation}
\begin{aligned}
    p_\data(\x)\left(\log\frac{p_\theta(\x)}{p_\data(\x)}\right)^2&\leq 2\max\left\{\frac{2+e^{-M}+e^M}{1+p_{\tref}(\x)/p_\data(\x)},\frac{p_\data(\x)+p_{\tref}(\x)}{p_{\tref}(\x)}\right\}(\Lc(\theta)_{\x}-\Lc(\theta^*)_{\x})\\
    &\leq 2\max\left\{\frac{2+e^{-M}+e^M}{1+e^{M_1}},1+e^{-M_1}\right\}(\Lc(\theta)_{\x}-\Lc(\theta^*)_{\x})
\end{aligned}
\end{equation}
Applying Jensen's inequality, we derive an upper bound for the forward KL divergence:
\begin{equation}
\begin{aligned}
    \kl{p_\data}{p_\theta}&=\E_{p_\data(\x)}\left[\log\frac{p_\data(\x)}{p_\theta(\x)}\right]\leq \sqrt{\E_{p_\data(\x)}\left[\left(\log\frac{p_\data(\x)}{p_\theta(\x)}\right)^2\right]}=\sqrt{\int p_\data(\x)\left(\log\frac{p_\theta(\x)}{p_\data(\x)}\right)^2\dm\x}\\
    &\leq C_1\sqrt{\Lc(\theta)-\Lc^*}
\end{aligned}
\end{equation}
where $C_1=\sqrt{2\max\left\{\frac{2+e^{-M}+e^M}{1+e^{M_1}},1+e^{-M_1}\right\}}$ is related to the lower bound of $\log\frac{p_{\tref}}{p_\data}$. Similarly, 
\begin{equation}
\begin{aligned}
    p_\theta(\x)\left(\log\frac{p_\theta(\x)}{p_\data(\x)}\right)^2&\leq 2\frac{p_\theta(\x)}{p_{\tref}(\x)}\max\left\{\frac{2+e^{-M}+e^M}{1+p_\data(\x)/p_{\tref}(\x)},\frac{p_\data(\x)+p_{\tref}(\x)}{p_{\data}(\x)}\right\}(\Lc(\theta)_{\x}-\Lc(\theta^*)_{\x})\\
    &\leq 2e^M\max\left\{\frac{2+e^{-M}+e^M}{1+e^{M_2}},1+e^{-M_2}\right\}(\Lc(\theta)_{\x}-\Lc(\theta^*)_{\x})
\end{aligned}
\end{equation}
By integrating over $\x$, we obtain $\kl{p_\theta}{p_\data}\leq C_2\sqrt{\Lc(\theta)-\Lc^*}$, where $C_2=\sqrt{2e^M\max\left\{\frac{2+e^{-M}+e^M}{1+e^{M_2}},1+e^{-M_2}\right\}}$ is related to the upper bound of $\log\frac{p_{\tref}}{p_\data}$.
\subsection{Analyses of $\Lc_{\alpha,\beta}(\theta)$}
Once we introduce additional coefficients $\alpha,\beta$, the generalized DDO objective $\Lc_{\alpha,\beta}(\theta)$ may become intractable and no longer admit $p_{\theta^*}=p_\data$ as the optimal solution. Specifically, the pointwise loss with $\alpha,\beta$ is
\begin{equation}
    \Lc_{\alpha,\beta}(\theta)_{\x}=-p_\data(\x)\log\sigma(\beta r_\theta(\x))-\alpha p_{\tref}(\x)\log(1-\sigma(\beta r_\theta(\x)))
\end{equation}
The optimal $\theta$ should satisfy
\begin{equation}
\begin{aligned}
    &\frac{\dm \Lc_{\alpha,\beta}(\theta)_{\x}}{\dm r_\theta(\x)}=\alpha\beta p_{\tref}(\x)\sigma(\beta r_\theta(\x))-\beta p_\data(\x)(1-\sigma(\beta r_\theta(\x)))=0\\
    \Rightarrow& \sigma(\beta r_\theta(\x))=\frac{p_\data(\x)}{p_\data(\x)+\alpha p_{\tref}(\x)}=\sigma\left(\log\frac{p_\data(\x)}{\alpha p_{\tref}(\x)}\right)\\
    \Rightarrow&p_\theta(\x)=p_{\tref}(\x)\left(\frac{p_\data(\x)}{\alpha p_{\tref}(\x)}\right)^{1/\beta}
\end{aligned}
\end{equation}
However, since $p_\theta$ is parameterized as a likelihood-based generative model, it must have self-normalized density. This optimality condition is only achieved when $\alpha$ is a proper normalizing constant satisfying
\begin{equation}
    \int p_\theta(\x)\dm\x=1\Rightarrow \alpha=\left(\int p_{\tref}^{1-1/\beta}(\x)p_\data^{1/\beta}(\x) \dm\x\right)^\beta
\end{equation}
Under this specific choice of $\alpha$, the optimal solution $p_{\theta^*}\propto p_{\tref}^{1-1/\beta}p_\data^{1/\beta}$. Otherwise, the optimization is subject to the constraint $\int p_\theta(\x)\dm\x=1$. To enforce this, we introduce a Lagrange multiplier $\lambda$ and define the Lagrangian:
\begin{equation}
    \Lc=\int \Lc_{\alpha,\beta}(\theta)_{\x}\dm\x+\lambda\left(1-\int p_\theta(\x)\dm\x\right)
\end{equation}
To find the optimal $p_\theta$, we take the functional derivative of $\Lc$ w.r.t. $p_\theta$ and set it to zero:
\begin{equation}
    \frac{\delta \Lc}{\delta p_\theta(\x)}=\frac{\dm \Lc_{\alpha,\beta}(\theta)_{\x}}{\dm p_\theta(\x)}-\lambda=\left(\alpha\beta p_{\tref}(\x)\sigma(\beta r_\theta(\x))-\beta p_\data(\x)(1-\sigma(\beta r_\theta(\x)))\right)\frac{\dm r_\theta(\x)}{\dm p_\theta(\x)}-\lambda=0
\end{equation}
which can be simplified to
\begin{equation}
    \alpha\left(\frac{p_\theta(\x)}{p_{\tref}(\x)}\right)^\beta-\frac{\lambda}{\beta}\frac{p_\theta(\x)}{p_{\tref}(\x)}\left[1+\left(\frac{p_\theta(\x)}{p_{\tref}(\x)}\right)^\beta\right]=\frac{p_\data(\x)}{p_{\tref}(\x)}
\end{equation}
This equation, combined with the constraint $\int p_\theta(\x)\dm\x=1$, determines both $\lambda$ and the optimal $p_\theta$. However, no closed-form solution exists for this problem. Despite this, we can expect certain ranges of $\alpha$ to skew the optimal solution away from $p_\data$ toward the direction of $p_{\tref}^{1-1/\beta}p_\data^{1/\beta}$, even if the exact equality does not hold.
\section{Experiment Details}
\label{appendix:exp-details}
Throughout all experiments, each training run under a given set of configurations (certain reference model and hyperparameters $\alpha,\beta$) is conducted on a single node with 8 NVIDIA A100 (SXM4-80GB) GPUs.

\paragraph{Diffusion Models}
We follow the parameterization and noise schedule of EDM~\citep{karras2022elucidating} and EDM2~\citep{karras2024analyzing}. Specifically, EDM introduces a time-dependent skip connection that preconditions the denoiser $\Dv_\theta$ (which predicts clean data $\x_0$) using a free-form network $\Fv_\theta$, allowing $\Fv_\theta$ to predict an adaptive mixture of signal and noise:
\begin{equation}
\Dv_\theta(\x_t,t)=c_\skipp(t)\x_t+c_\out(t)\Fv_\theta(c_\inn(t)\x_t,c_\noise(t))
\end{equation}
where
\begin{equation}
    c_{\skipp}(t)=\frac{\sigma_\data^2}{\sigma_\data^2+t^2},\quad c_{\out}(t)=\frac{\sigma_\data t}{\sqrt{\sigma_\data^2+t^2}},\quad c_{\inn}(t)=\frac{1}{\sqrt{\sigma_\data^2+t^2}},\quad c_\noise(t)=\frac{1}{4}\log t
\end{equation}
EDM employs a simple variance exploding (VE) noise schedule satisfying $\alpha_t=1,\sigma_t=t$. It's worth noting that the preconditioning used in EDM actually transforms it into v-prediction under the variance preserving (VP) noise schedule, owing to the normalizing factor $c_{\inn}(t)$ and the skip connection coefficients $c_{\skipp}(t), c_{\out}(t)$~\citep{zheng2023improved}. The EDM models are pretrained by a F-prediction MSE loss:
\begin{equation}
    \Lc^{\text{EDM}}(\theta)=\E_{\x_0\sim p_\data,t\sim p(t),\epsilonv\sim\Nc(\vect 0,\Iv)}\left[\|\Fv_\theta(c_{\inn}(t)\x_t,c_\noise(t))-\hat\Fv(\x_0,\x_t,t)\|_2^2\right]
\end{equation}
where $\x_t=\x_0+t\epsilonv, \hat\Fv(\x_0,\x_t,t)=\frac{\x_0-c_\skipp(t)\x_t}{c_\out(t)}$ is the prediction target, and $p(t)$ is a time distribution satisfying $\log t\sim\Nc(P_{\text{mean}},P_{\text{std}}^2)$, where $P_{\text{mean}},P_{\text{std}}$ are hyperparameters.

We adopt similar settings for DDO finetuning. Specifically, we use the approximation in Eqn.~\eqref{eq:diffusion-ddo-approx} and set the weighting $w(t)$ to satisfy $w(t)\|\epsilonv_\theta-\epsilonv\|_2^2=\|\Fv_\theta-\hat\Fv\|_2^2$, leading to the following objective ($1-\sigma(x)=\sigma(-x)$):
\begin{equation}
\begin{aligned}
    \Lc^{\text{EDM-DDO}}_{\alpha,\beta}(\theta)=-\E_{t\sim p(t),\epsilonv\sim\Nc(\vect 0,\Iv)}\Big[&\E_{p_\data(\x_0)}\log \sigma\left(-\beta\left(\|\Fv_\theta-\hat\Fv\|_2^2-\|\Fv_{\tref}-\hat\Fv\|_2^2\right)\right)\\
    +\alpha&\E_{p_{\tref}(\x_0)}\log \sigma\left(\beta\left(\|\Fv_\theta-\hat\Fv\|_2^2-\|\Fv_{\tref}-\hat\Fv\|_2^2\right)\right)\Big]
\end{aligned}
\end{equation}
where we use the same form of time distribution $\log t\sim\Nc(P_{\text{mean}},P_{\text{std}}^2)$, and $P_{\text{mean}},P_{\text{std}}$ are typically the same as pertaining. See Appendix~\ref{appendix:example} for the specific implementation. For each finetuning round, we launch $\sim$20 nodes to sweep over the hyperparameters $\alpha,\beta$ in $[0.5, 6.0]\times [0.01,0.1]$. We disable all dropout layers in the network to ensure steady improvement. We also find numerical precision crucial for the diffusion DDO loss and disable mixed-precision training. For each round, 50k images are generated offline from the reference model as the reference dataset.

For EDM on CIFAR-10, we finetune the unconditional and class-conditional model for 12 and 16 rounds, respectively. We set $P_{\text{mean}}=-1.2,P_{\text{std}}=1.2$ throughout all rounds, which is the same as pretraining. Each round has a duration of 1.5M images (30 epochs, 0.75\% of pretraining) with a batch size of 512. The learning rate warms up linearly from 0 to $1.5e-4$ during each round, and the data augmentation probability is set to 12\% as pretraining. We find the exponential moving average (EMA) beneficial for stabilizing the model performance and choose a relatively small EMA half-life (0.25M images) as we finetune less duration than pretraining. We evaluate the FID each time trained with 50k images and save the best model for the next round. Each round takes $\sim$3h including both training and evaluation.

For EDM2-S on ImageNet-64, we finetune the model for 24 rounds, where we set $P_{\text{mean}}=-0.8,P_{\text{std}}=1.6$ for the first 16 rounds following pretraining, and increase $P_{\text{std}}$ to 3.0 in the last 8 rounds. Each round has a duration of 6.4M images (5 epochs, 0.6\% of pretraining) with a batch size of 512. We use a learning rate of $5e-5$ for the first 16 rounds and $2e-5$ for the last 8 rounds, along with the learning rate scheduler in EDM2 which is a mix of linear warmup and inverse square root decay, where we set the ramp-up to 1M images and the decay reference to 2000 iterations. We follow the power function EMA introduced in EDM2 and set the EMA length to 0.05. We evaluate the FID each time trained with $2^{17}$ ($\approx$ 131k) images and save the best model for the next round. Each round takes $\sim$1d including both training and evaluation.

For EDM2-L on ImageNet 512$\times$512, we finetune the model for 28 rounds. Following the pretraining setup, we fix $P_{\text{mean}} = -0.4$, while progressively increasing the variance $P_{\text{std}}$ to 1.6, 2.0, and 3.0 starting from the 1st, 18th, and 22nd round, respectively. Each round has a duration of 6.4M images (5 epochs, 0.34\% of pretraining) with a batch size of 2048. We use a learning rate of $1e-4$ for the first 24 rounds and $5e-5$ for the last 4 rounds, along with the learning rate scheduler in EDM2 which is a mix of linear warmup and inverse square root decay, where we set the ramp-up to 1M images and the decay reference to 800 iterations. We follow the power function EMA introduced in EDM2 and set the EMA length to 0.05. We evaluate the FID each time trained with $2^{18}$ ($\approx$ 262k) images and save the best model for the next round. Each round takes $\sim$2d including both training and evaluation. We further boost the finetuned model with autoguidance~\citep{karras2024guiding}. We choose \texttt{edm2-img512-xs-0134217-0.165.pkl} from~\citet{karras2024guiding} as the bad version model and use a small guidance scale of 1.1, which slightly improves the FID from 1.26 to 1.21.

\paragraph{Autoregressive Models} We finetune VAR-d16 and VAR-d30~\citep{tian2024visual} both for only 2 rounds. VAR is highly efficient at inference, enabling us to sample from the reference distribution online during training by generating random latent tokens with the reference model conditioned on the same class labels as those in the dataset batch. We disable all dropout layers in the network and set the label dropout probability to 50\% for unconditional training. Unlike diffusion DDO, we enable mixed-precision when finetuning VAR. For each round, we launch $\sim$10 nodes to sweep over the hyperparameters $\alpha,\beta$ in $[10.0,100.0]\times\{0.02\}$. Each round has a duration of 80 iterations (0.064 epoch, less than 0.03\% of pretraining) with a batch size of 1024. We follow the learning rate scheduler in VAR pretraining and set the peak learning rate to a smaller value of $4e-6$. We evaluate the FIDs (corresponding to guidance-free/a moderate CFG scale) every 4 iterations and save the best model for the next round. Each round takes $\sim$5h/7.5h for VAR-d16/d30 including both training and evaluation.
\section{Code Example}
\label{appendix:example}
\begin{minted}[fontsize=\small, bgcolor=white]{python}
import torch
import torch.nn.functional as F


# Original diffusion loss of EDM
class EDMLoss:
    def __init__(self, P_mean=-0.4, P_std=1.0, sigma_data=0.5):
        self.P_mean = P_mean
        self.P_std = P_std
        self.sigma_data = sigma_data

    def __call__(self, net, images, labels=None):
        """
        net: the target denoiser network
        images: real samples from the dataset, shape (B, C, H, W)
        """
        # Sample diffusion time
        rnd_normal = torch.randn([images.shape[0], 1, 1, 1], device=images.device)
        sigma = (rnd_normal * self.P_std + self.P_mean).exp()
        # Diffusion loss weighting
        weight = (sigma**2 + self.sigma_data**2) / (sigma * self.sigma_data) ** 2
        # Sample Gaussian noise
        noise = torch.randn_like(images) * sigma
        # Denoise
        denoised = net(images + noise, sigma, labels)
        # Compute loss
        loss = torch.sum(weight * (denoised - images) ** 2, dim=(1, 2, 3))
        return loss.mean()


# Diffusion DDO loss of EDM
class EDMLoss_DDO:
    def __init__(self, P_mean=-0.4, P_std=1.0, sigma_data=0.5, alpha=1.0, beta=0.02):
        self.P_mean = P_mean
        self.P_std = P_std
        self.sigma_data = sigma_data
        self.alpha = alpha
        self.beta = beta

    def __call__(self, net, ref_net, images, fake_images, labels=None, fake_labels=None):
        """
        net: the target denoiser network
        ref_net: the reference denoiser network (frozen)
        images: real samples from the dataset, shape (B, C, H, W)
        fake_images: fake samples generated by the reference model, shape (B, C, H, W)
        """
        # Sample diffusion time
        rnd_normal = torch.randn([images.shape[0], 1, 1, 1], device=images.device)
        sigma = (rnd_normal * self.P_std + self.P_mean).exp()
        # Diffusion loss weighting
        weight = (sigma**2 + self.sigma_data**2) / (sigma * self.sigma_data) ** 2
        # Sample Gaussian noise
        noise = torch.randn_like(images) * sigma
        # Denoise by the target model
        D = net(images + noise, sigma, labels)
        net.eval()
        D_fake = net(fake_images + noise, sigma, fake_labels)
        net.train()
        D_logp = -torch.sum(weight * (D - images) ** 2, dim=(1, 2, 3))
        D_fake_logp = -torch.sum(weight * (D_fake - fake_images) ** 2, dim=(1, 2, 3))
        # Denoise by the reference model
        with torch.no_grad():
            ref_D = ref_net(images + noise, sigma, labels)
            ref_D_fake = ref_net(fake_images + noise, sigma, fake_labels)
        ref_D_logp = -torch.sum(weight * (ref_D - images) ** 2, dim=(1, 2, 3))
        ref_D_fake_logp = -torch.sum(weight * (ref_D_fake - fake_images) ** 2, dim=(1, 2, 3))
        # Compute loss
        loss = -F.logsigmoid(self.beta * (D_logp - ref_D_logp)) \
               -self.alpha * F.logsigmoid(-self.beta * (D_fake_logp - ref_D_fake_logp))
        return loss.mean()

\end{minted}

\section{Additional Results}
\label{appendix:additional-results}

\newpage
\clearpage

\begin{figure}[!ht]
	\centering		
 	\includegraphics[width=0.9\linewidth]{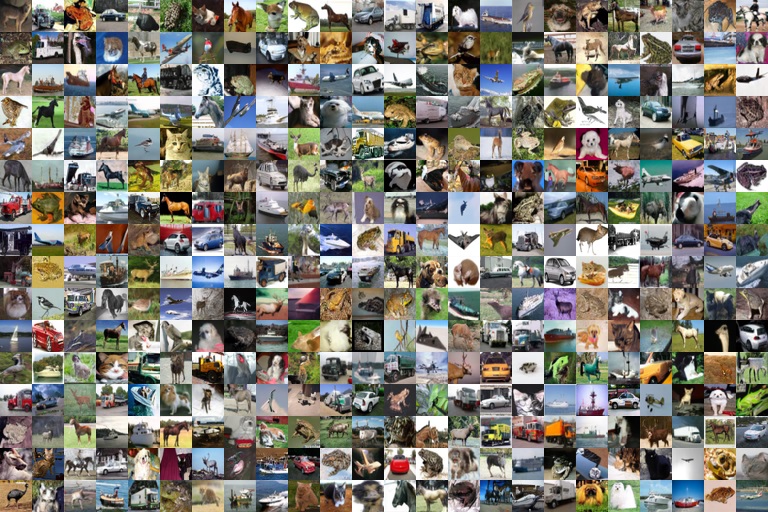}\\
 	\vspace{-.1in}
	\caption{Random samples of EDM (CIFAR-10, Unconditional), FID 1.97.}
 	\vspace{-.1in}
\end{figure}

\begin{figure}[!ht]
	\centering		
 	\includegraphics[width=0.9\linewidth]{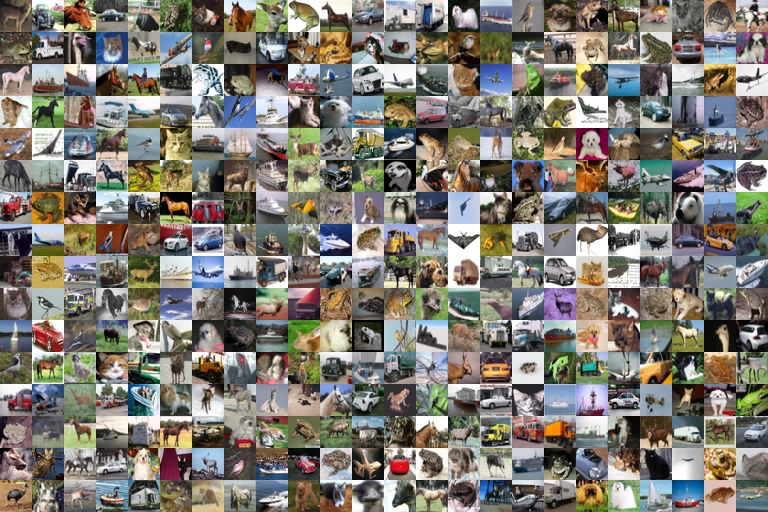}\\
 	\vspace{-.1in}
	\caption{Random samples of EDM + DDO (CIFAR-10, Unconditional), FID 1.38.}
 	\vspace{-.1in}
\end{figure}

\begin{figure}[!ht]
	\centering		
 	\includegraphics[width=0.9\linewidth]{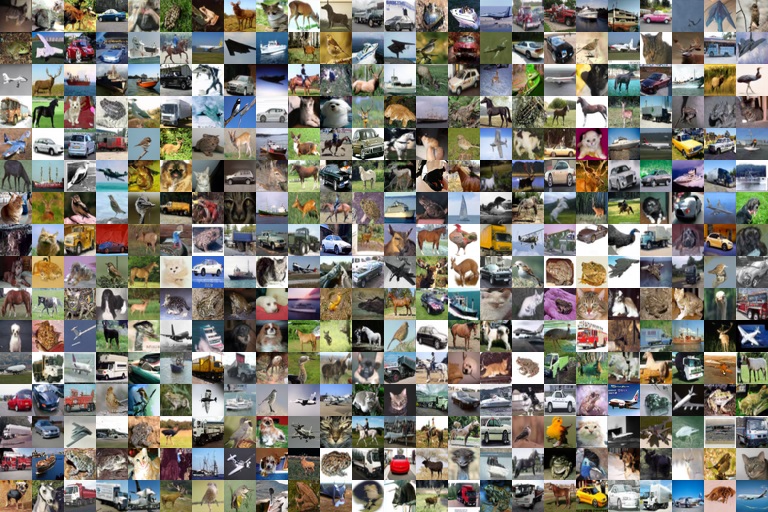}\\
 	\vspace{-.1in}
	\caption{Random samples of EDM (CIFAR-10, Class-conditional), FID 1.85.}
 	\vspace{-.1in}
\end{figure}

\begin{figure}[!ht]
	\centering		
 	\includegraphics[width=0.9\linewidth]{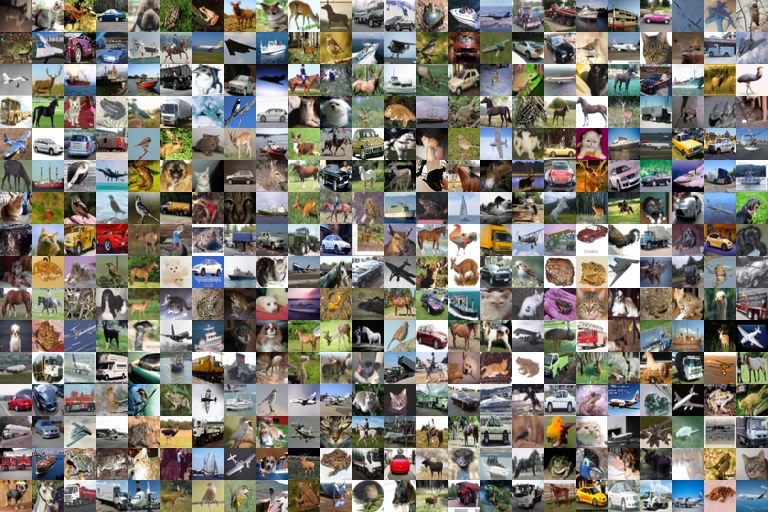}\\
 	\vspace{-.1in}
	\caption{Random samples of EDM + DDO (CIFAR-10, Class-conditional), FID 1.30.}
 	\vspace{-.1in}
\end{figure}

\begin{figure}[!ht]
	\centering		
 	\includegraphics[width=0.8\linewidth]{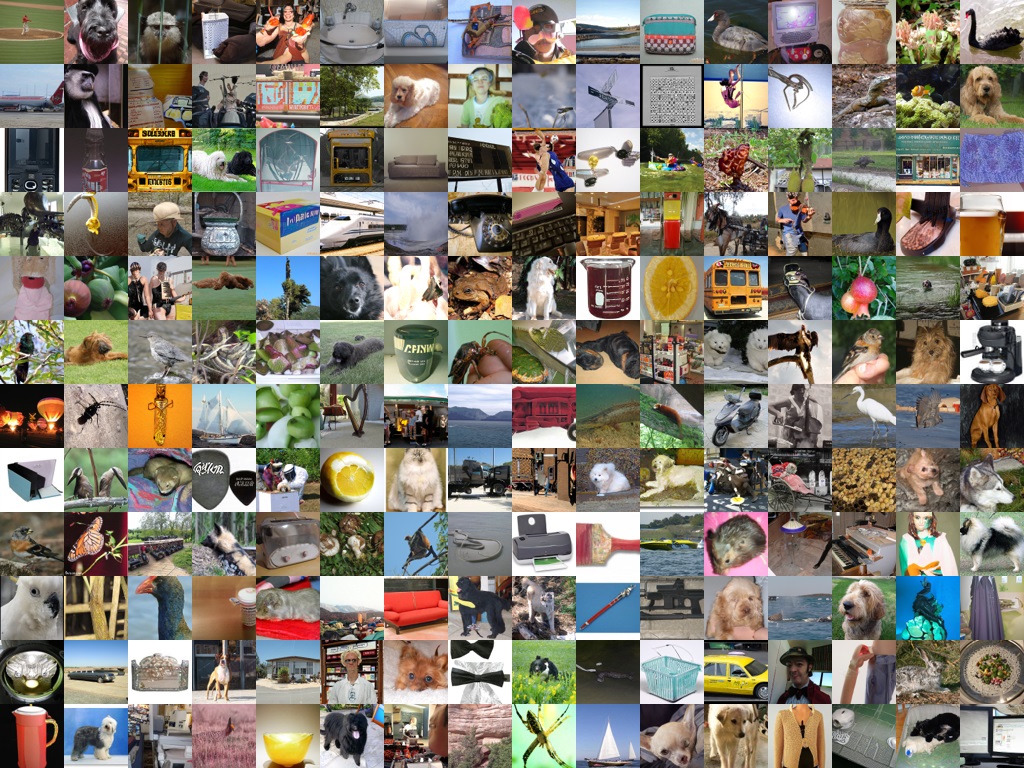}\\
 	\vspace{-.1in}
	\caption{Random samples of EDM2-S (ImageNet-64, Class-conditional), FID 1.60.}
 	\vspace{-.1in}
\end{figure}

\begin{figure}[!ht]
	\centering		
 	\includegraphics[width=0.8\linewidth]{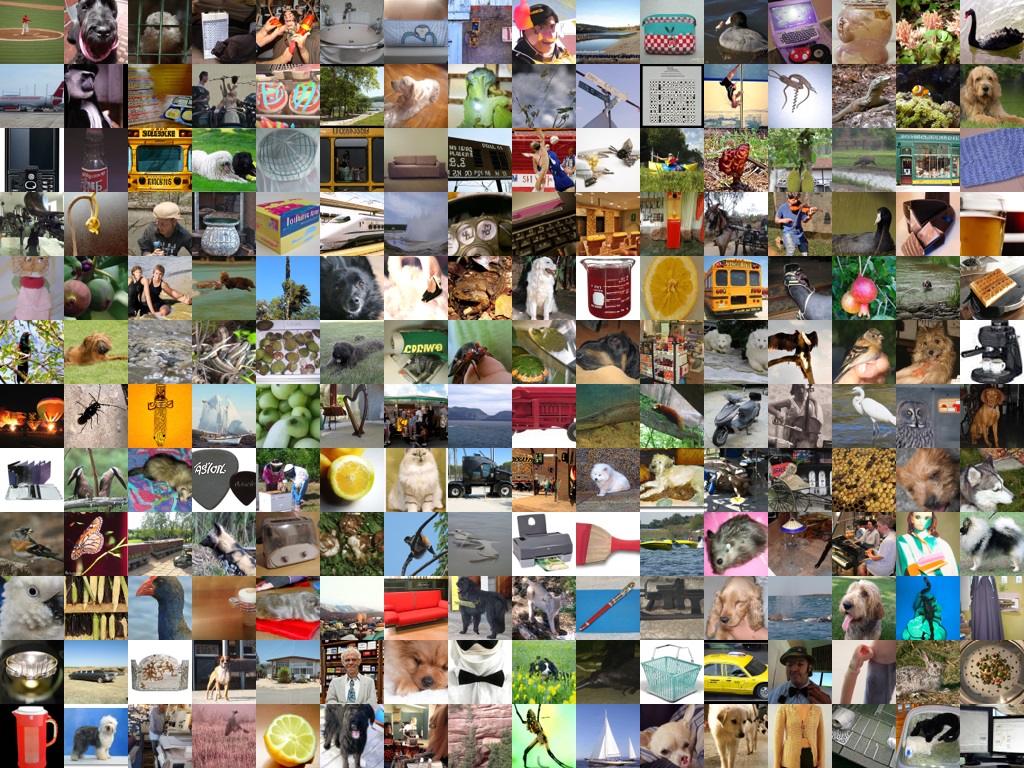}\\
 	\vspace{-.1in}
	\caption{Random samples of EDM2-S + DDO (ImageNet-64, Class-conditional), FID 0.97.}
 	\vspace{-.1in}
\end{figure}

\begin{figure}[ht]
	\centering		
 	\includegraphics[width=0.88\linewidth]{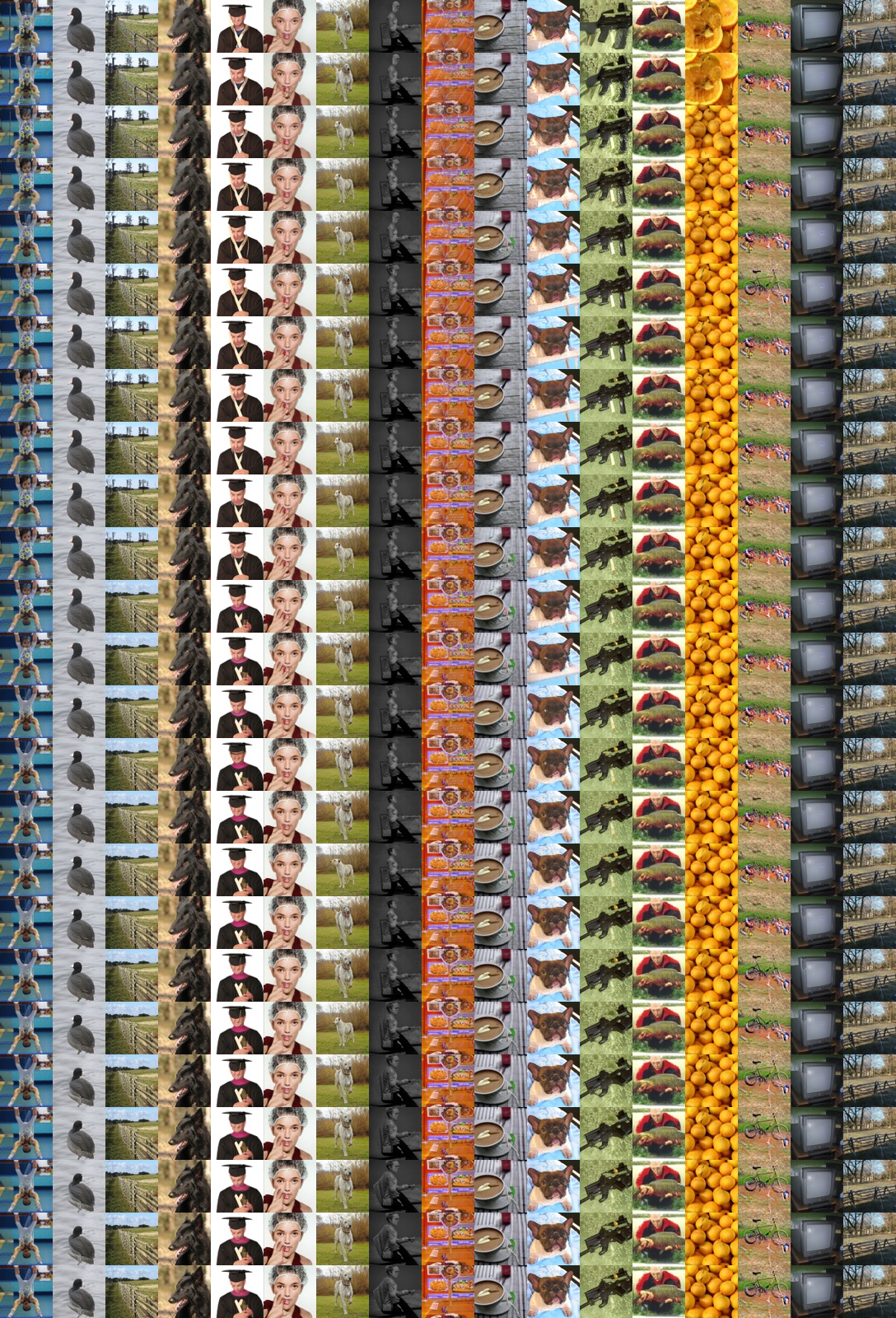}\\
 	\vspace{-.1in}
	\caption{Illustration of the multi-round refinement process on EDM2-S (ImageNet-64).}
 	\vspace{-.1in}
\end{figure}

\begin{figure}[ht]
    \vspace{-.1in}
    \centering
    \resizebox{\textwidth}{!}{
    \begin{tabular}{cc}
    \makecell{VAR-d16 w/o trick\\(FID 11.33)}&\includegraphics[width=\linewidth,align=c]{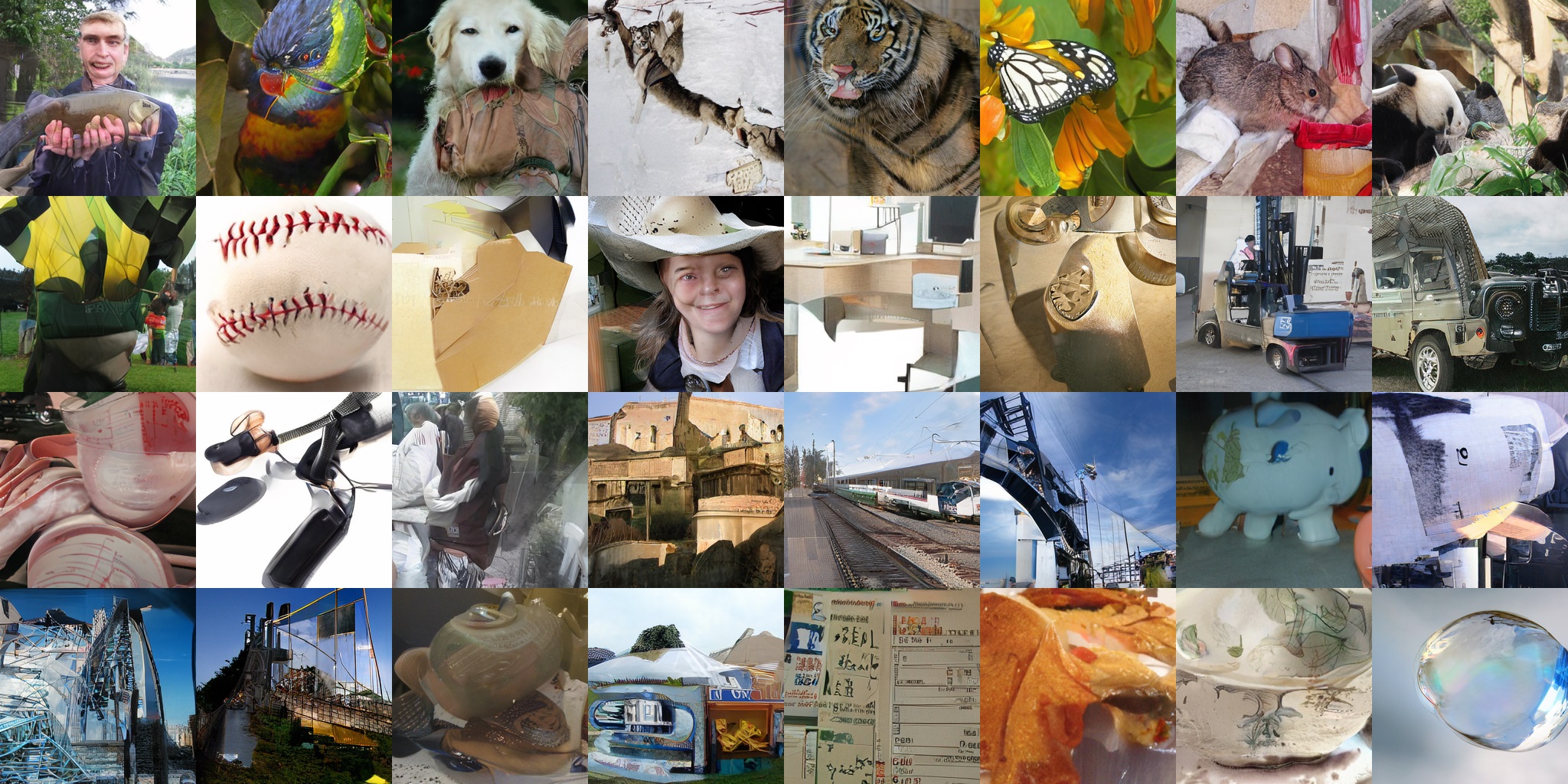}\\
    \\
    \makecell{VAR-d16 w/ trick\\(FID 3.62)}&\includegraphics[width=\linewidth,align=c]{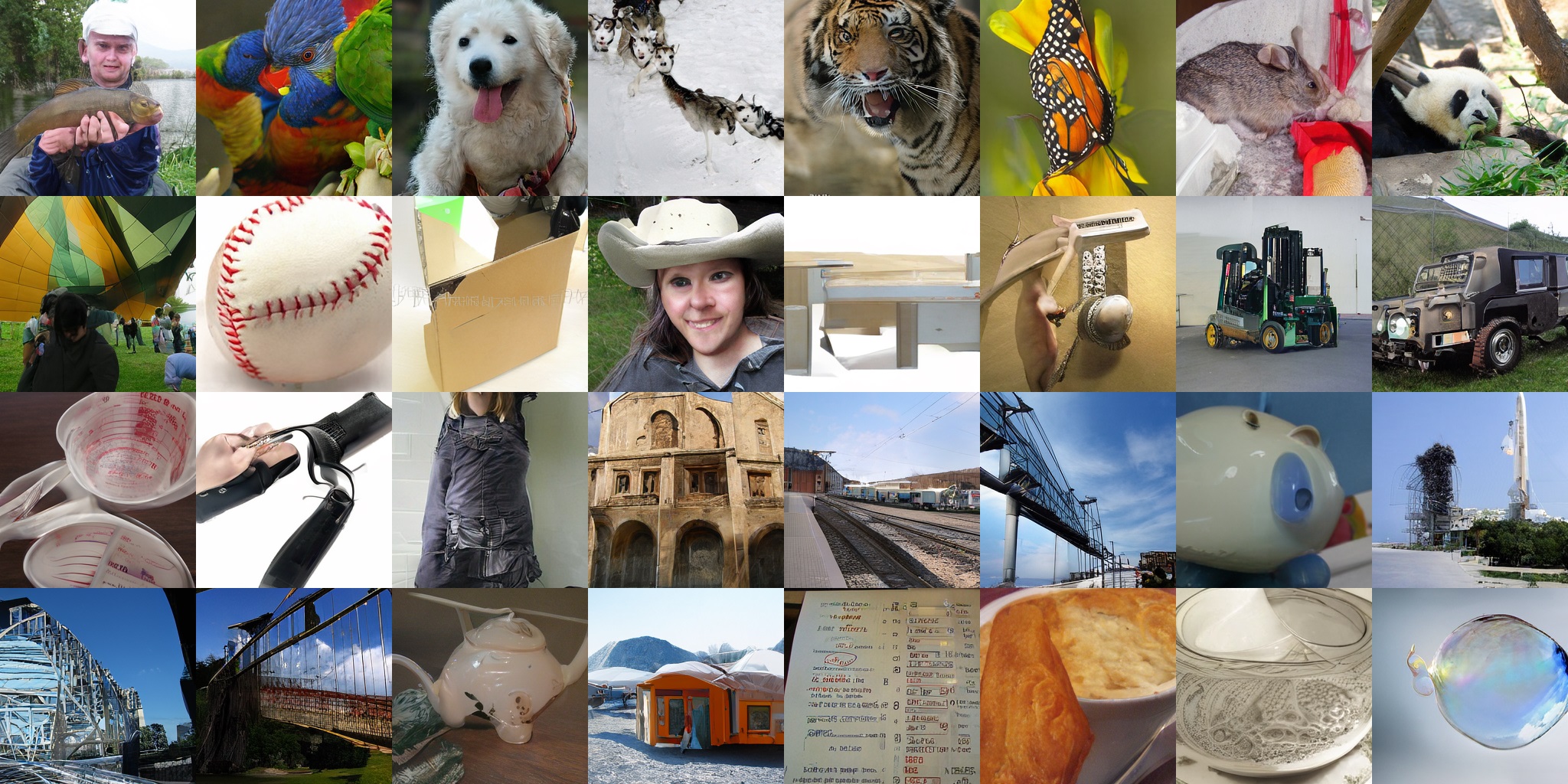}\\
    \\
    \makecell{VAR-d16 + DDO\\(FID 3.12)}&\includegraphics[width=\linewidth,align=c]{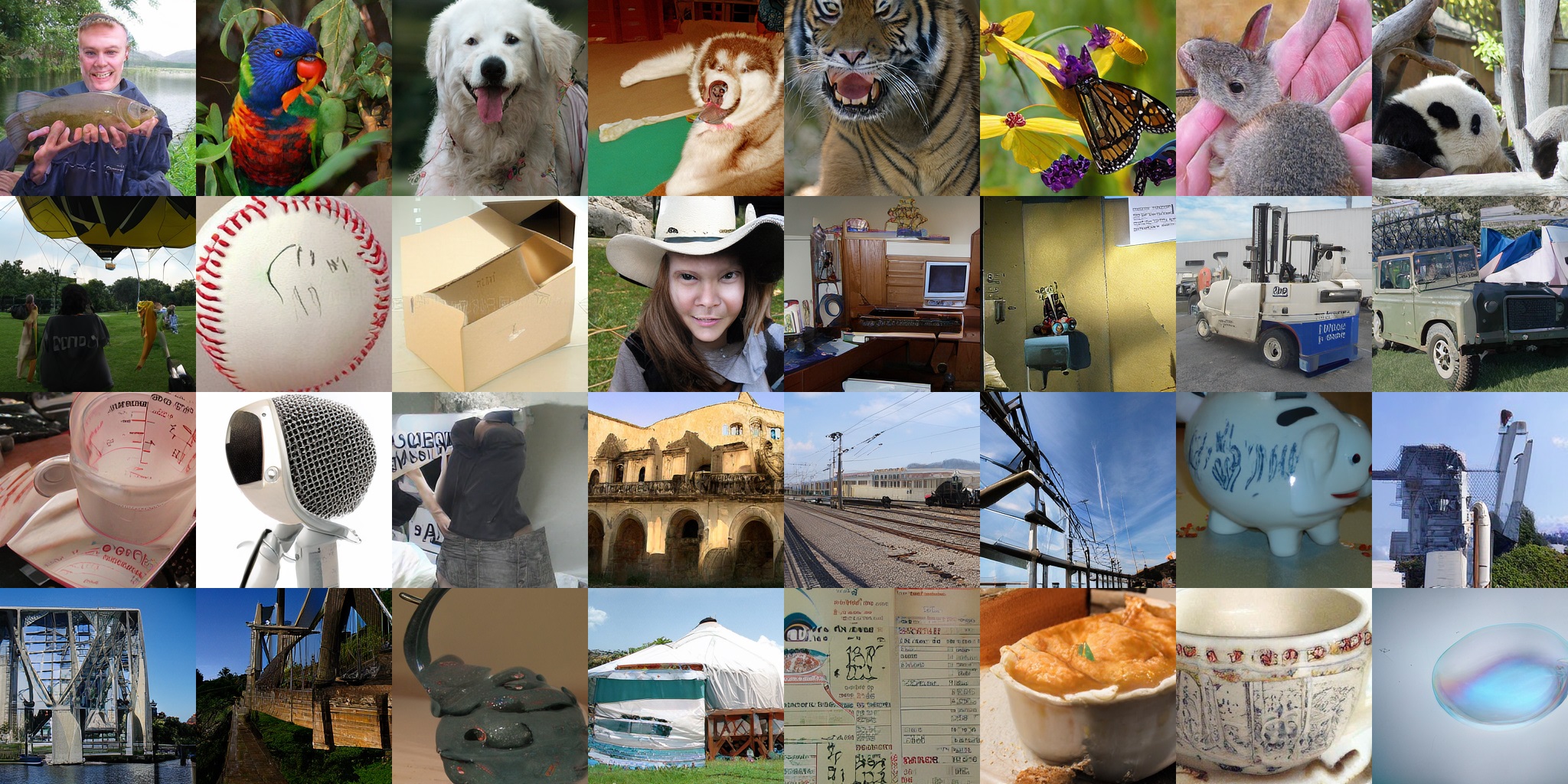}
    \end{tabular}
    }
    \caption{Guidance-free samples by pretrained and finetuned VAR-d16.}
    \vspace{-0.1in}
\end{figure}

\begin{figure}[ht]
    \vspace{-.1in}
    \centering
    \resizebox{\textwidth}{!}{
    \begin{tabular}{cc}
    \makecell{VAR-d16 w/o trick\\(FID 3.71)}&\includegraphics[width=\linewidth,align=c]{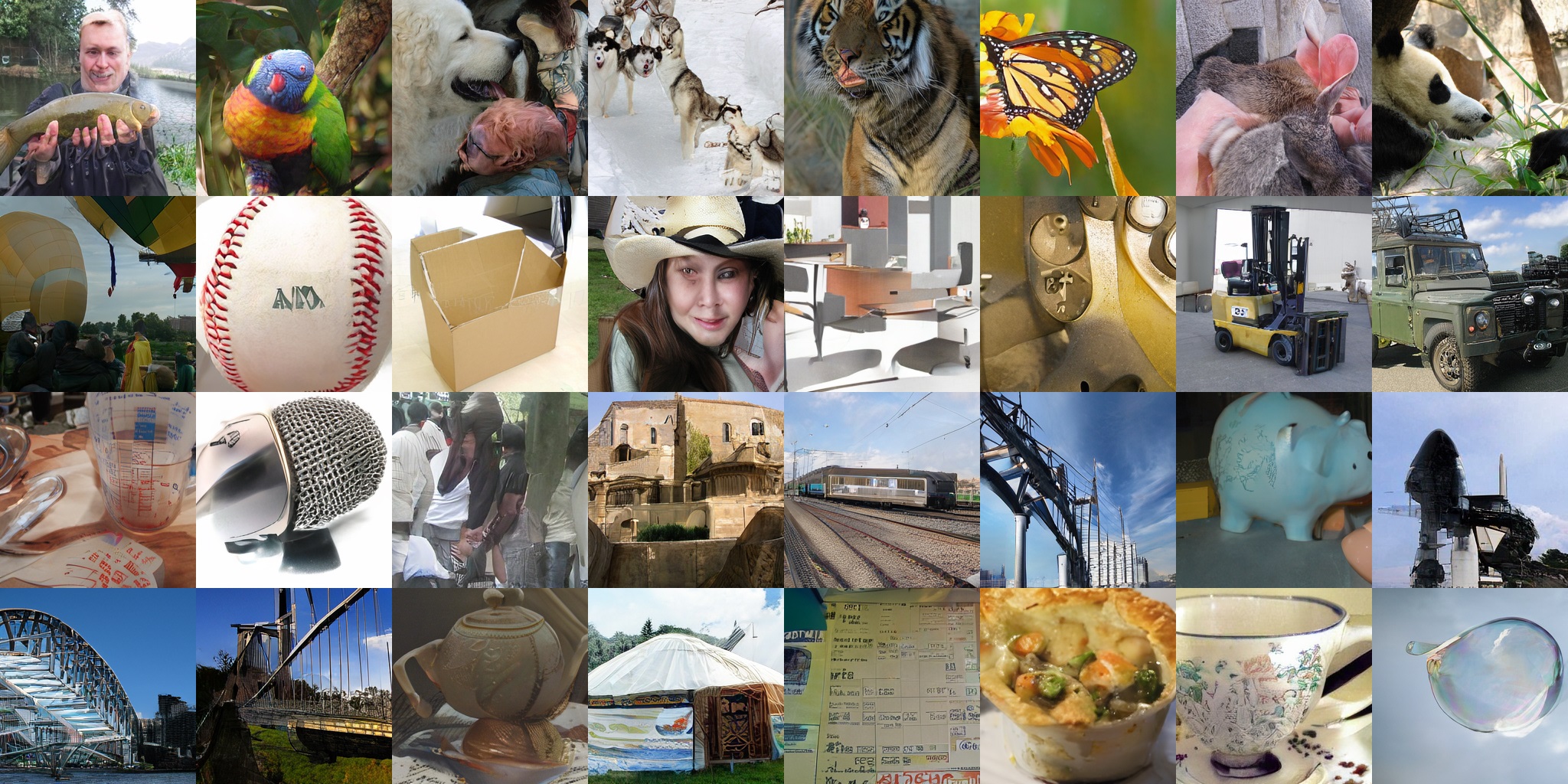}\\
    \\
    \makecell{VAR-d16 w/ trick\\(FID 3.30)}&\includegraphics[width=\linewidth,align=c]{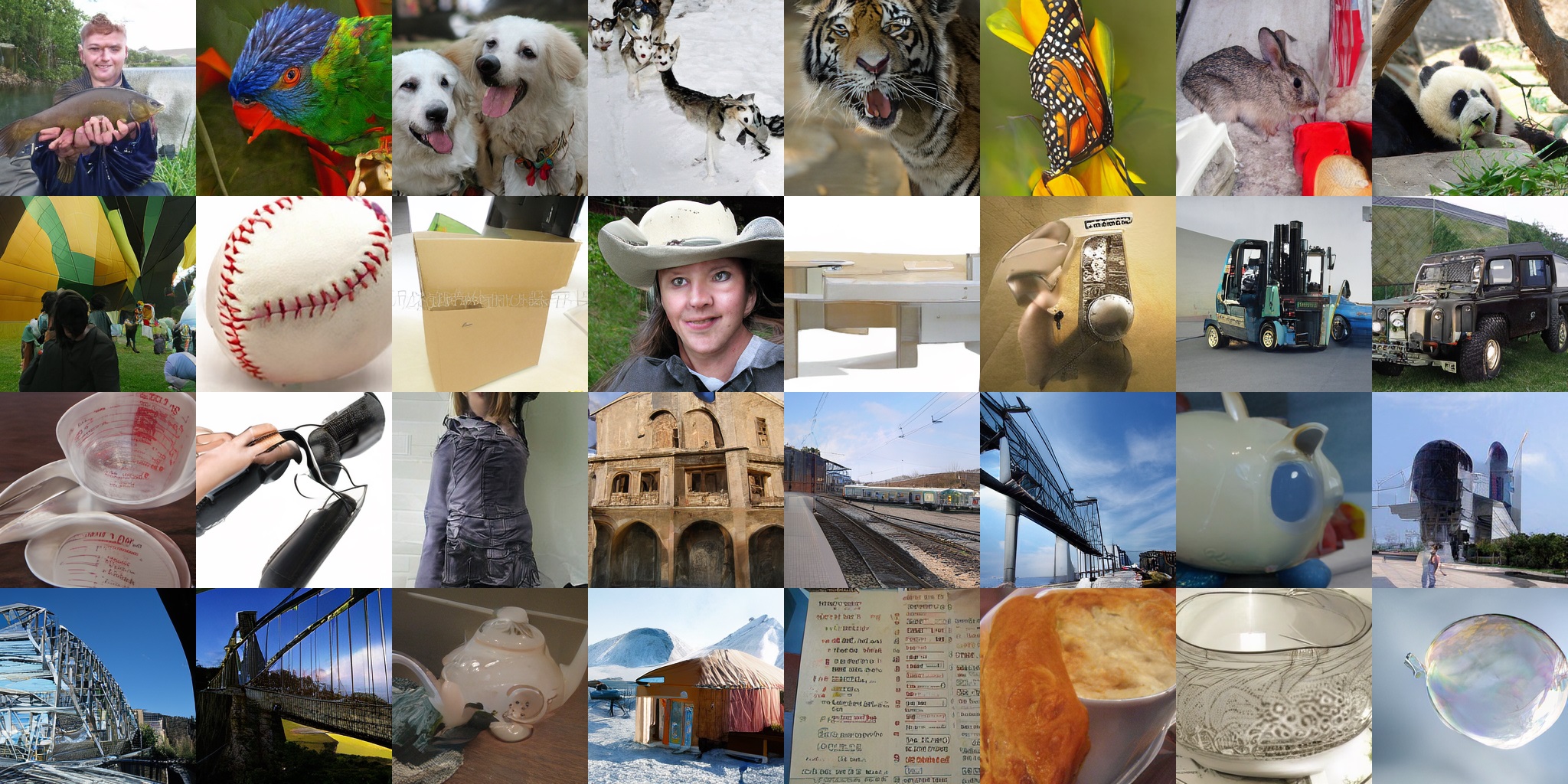}\\
    \\
    \makecell{VAR-d16 + DDO\\(FID 2.54)}&\includegraphics[width=\linewidth,align=c]{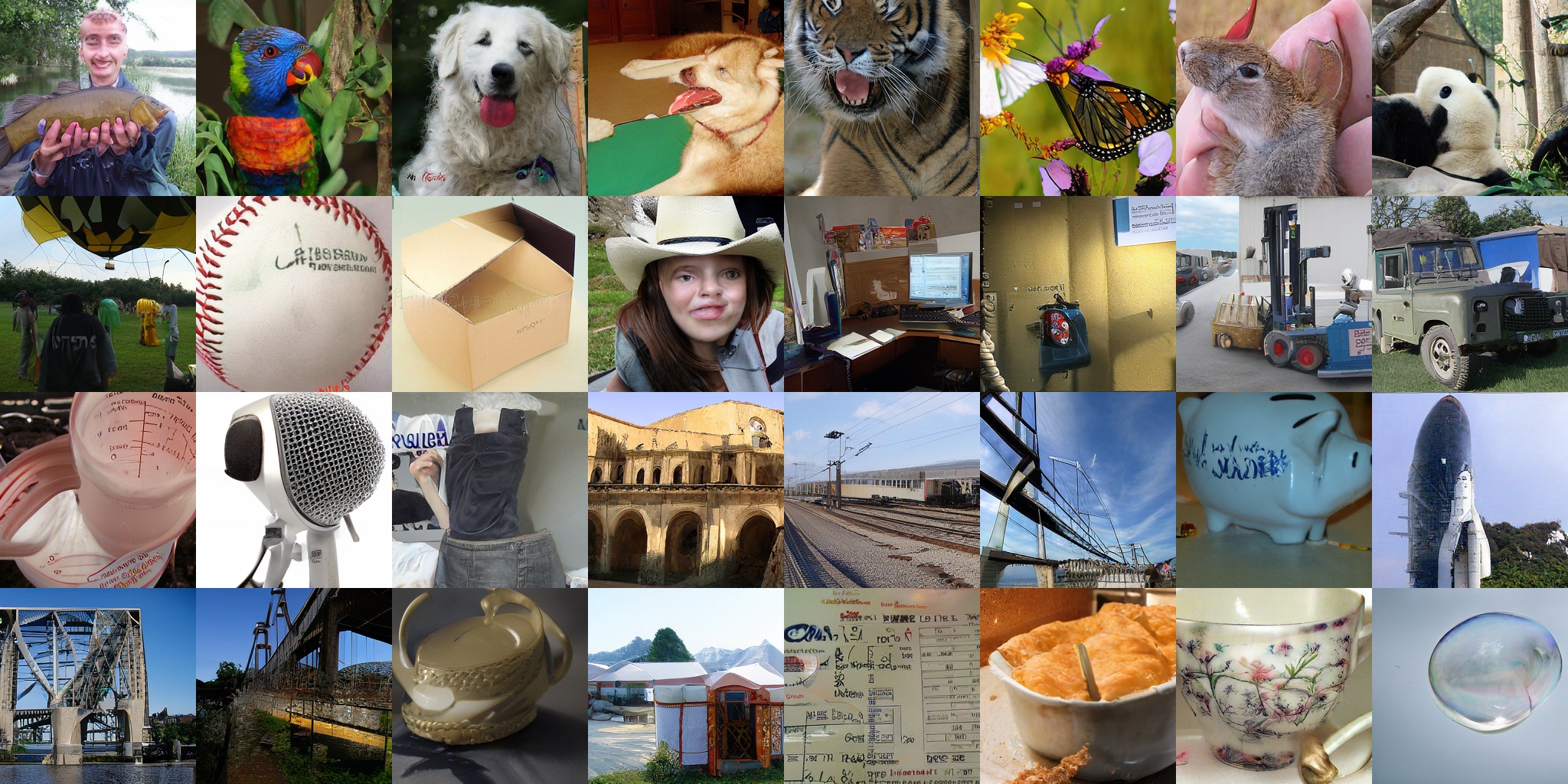}
    \end{tabular}
    }
    \caption{CFG-enhanced samples by pretrained and finetuned VAR-d16.}
    \vspace{-0.1in}
\end{figure}

\begin{figure}[ht]
    \vspace{-.1in}
    \centering
    \resizebox{\textwidth}{!}{
    \begin{tabular}{cc}
    \makecell{VAR-d30 w/o trick\\(FID 4.74)}&\includegraphics[width=\linewidth,align=c]{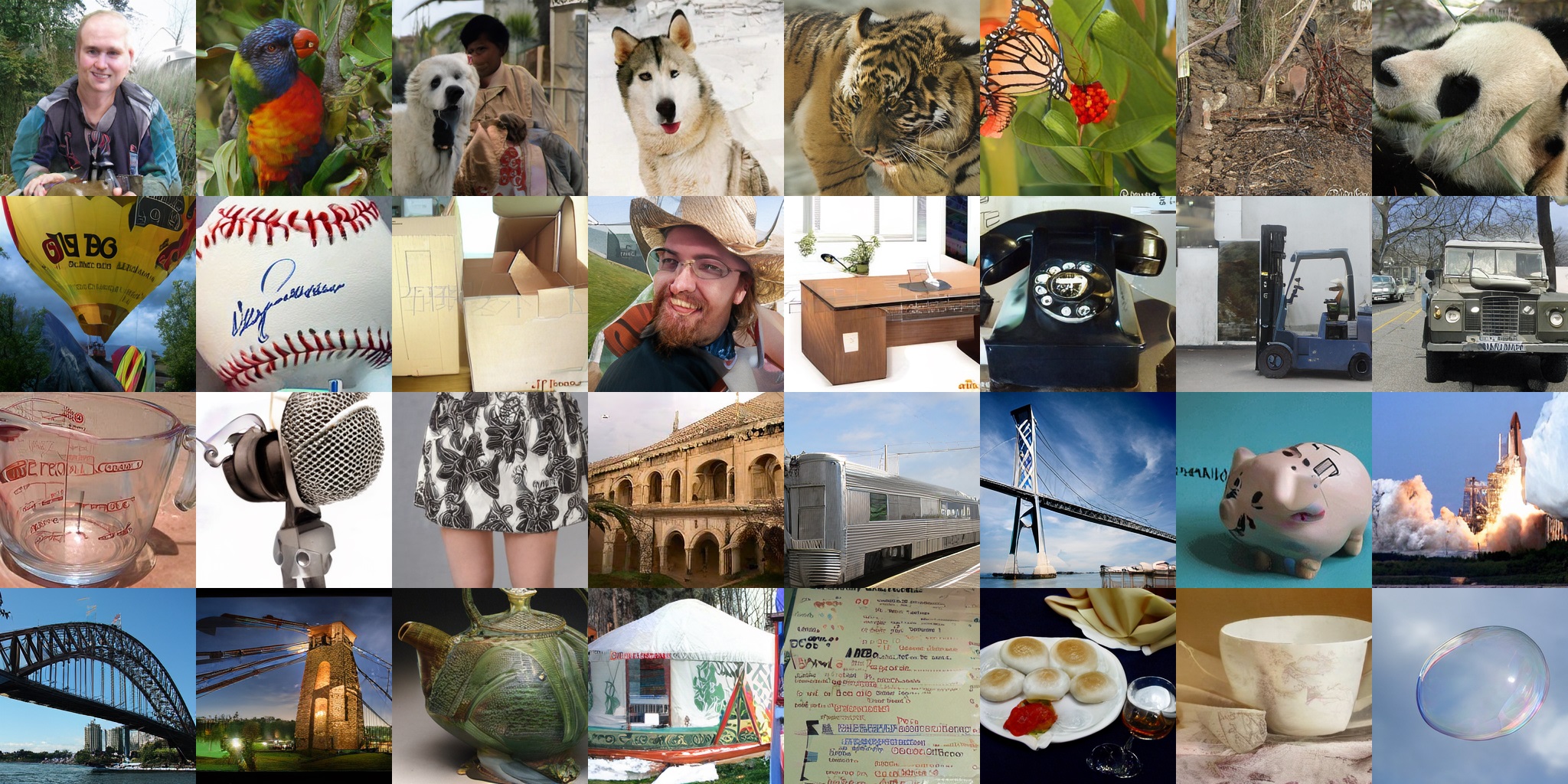}\\
    \\
    \makecell{VAR-d30 w/ trick\\(FID 2.17)}&\includegraphics[width=\linewidth,align=c]{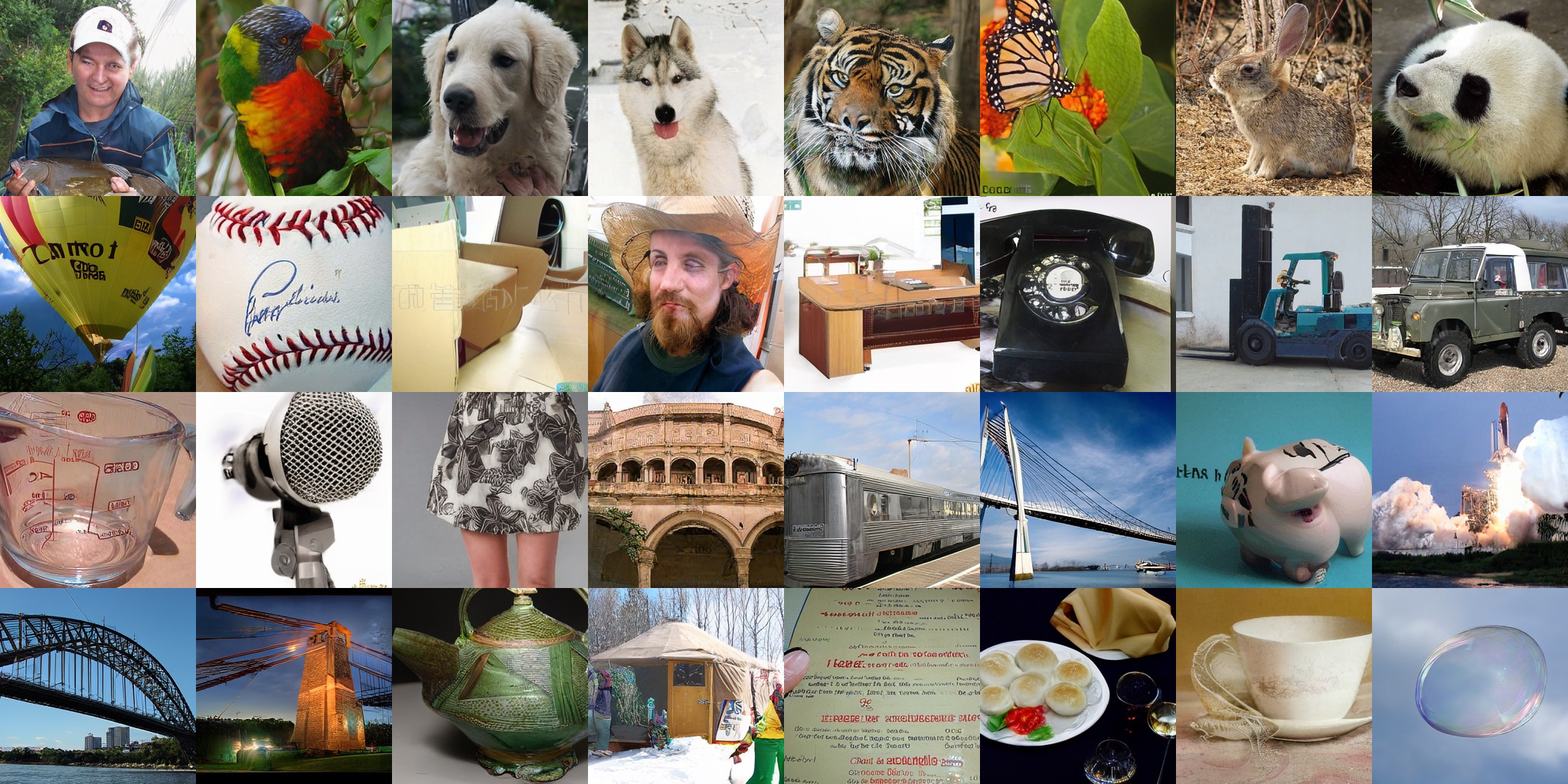}\\
    \\
    \makecell{VAR-d30 + DDO\\(FID 1.79)}&\includegraphics[width=\linewidth,align=c]{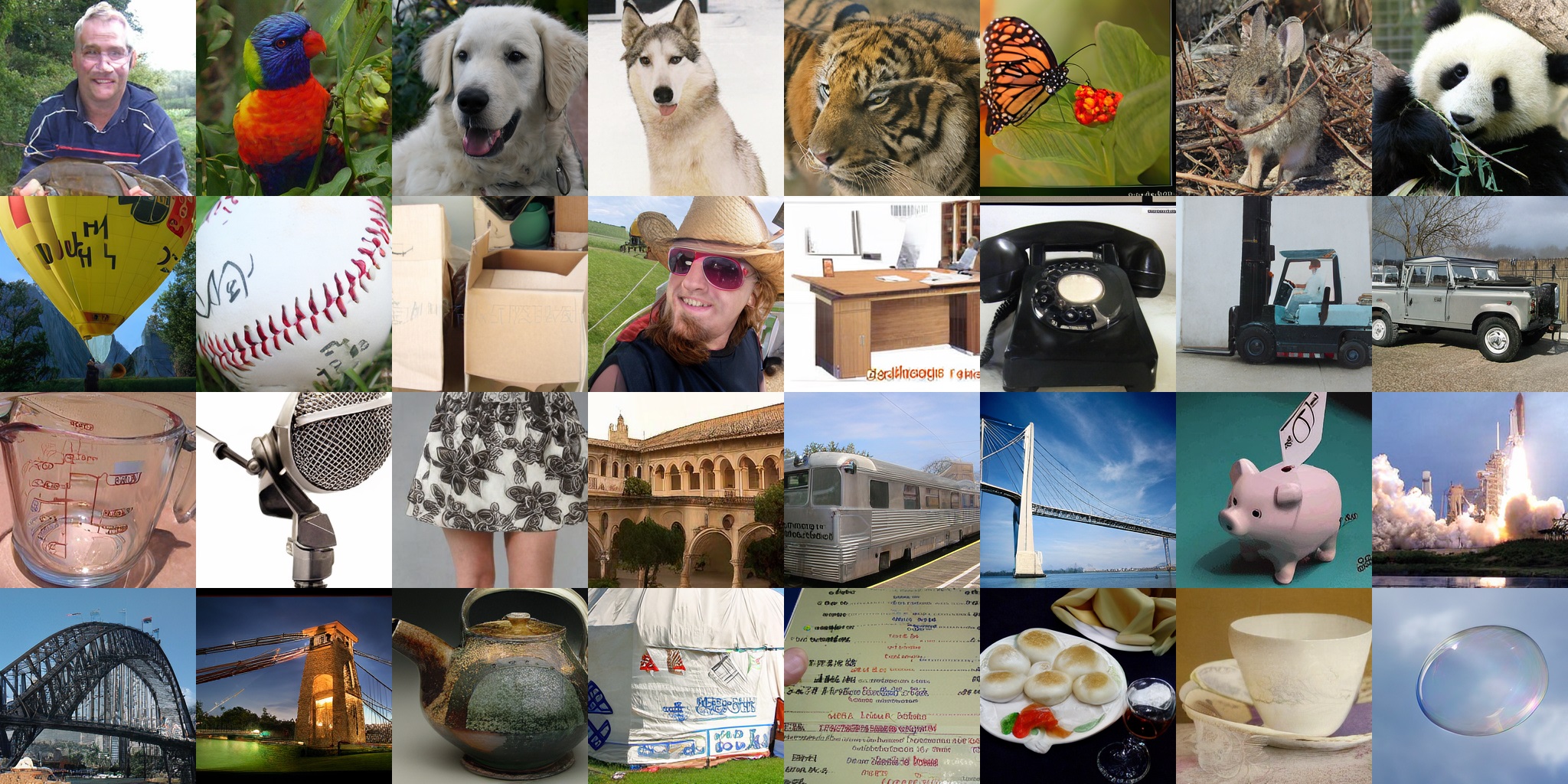}
    \end{tabular}
    }
    \caption{Guidance-free samples by pretrained and finetuned VAR-d30.}
    \vspace{-0.1in}
\end{figure}

\begin{figure}[ht]
    \vspace{-.1in}
    \centering
    \resizebox{\textwidth}{!}{
    \begin{tabular}{cc}
    \makecell{VAR-d30 w/o trick\\(FID 1.92)}&\includegraphics[width=\linewidth,align=c]{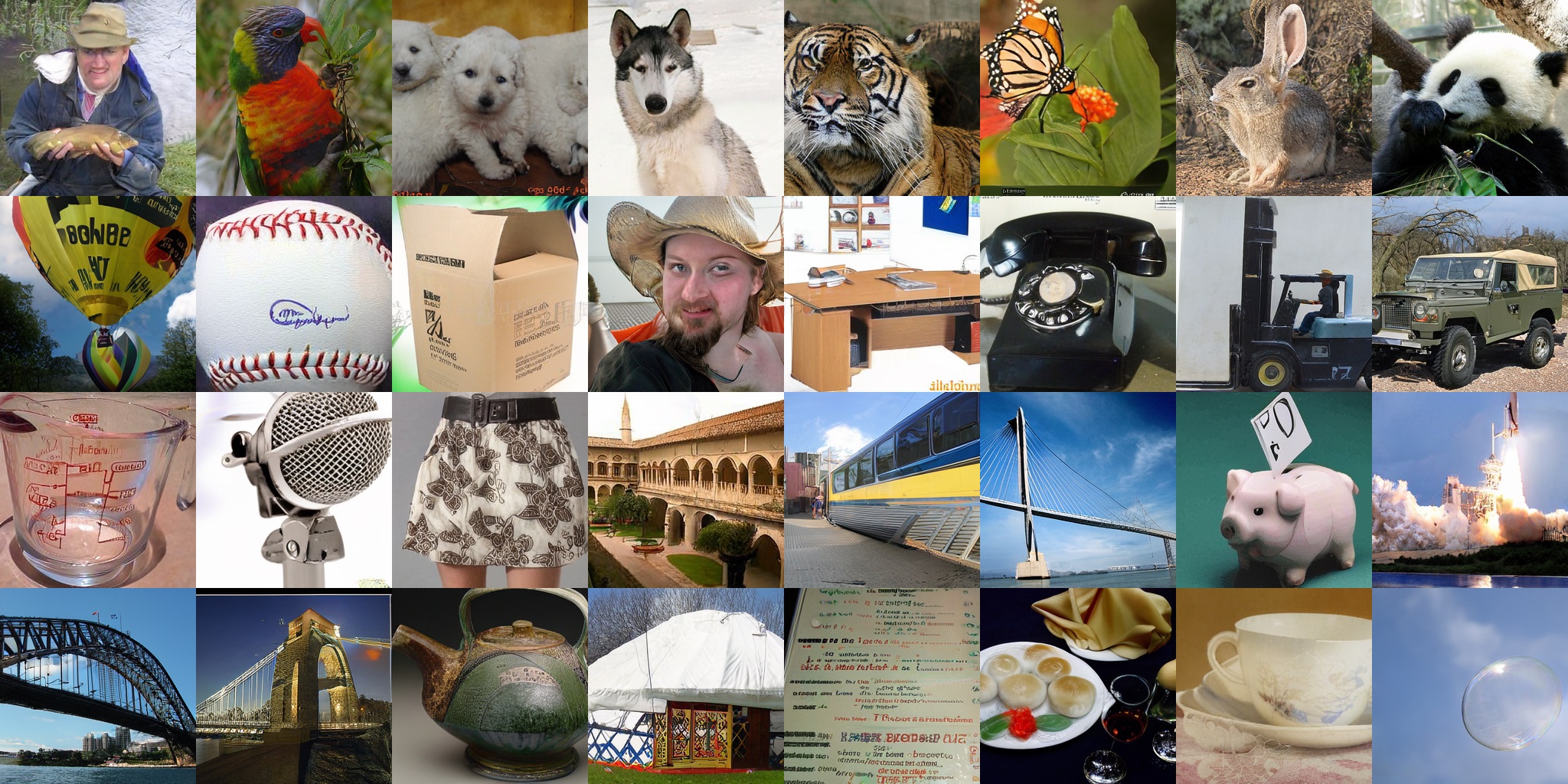}\\
    \\
    \makecell{VAR-d30 w/ trick\\(FID 1.90)}&\includegraphics[width=\linewidth,align=c]{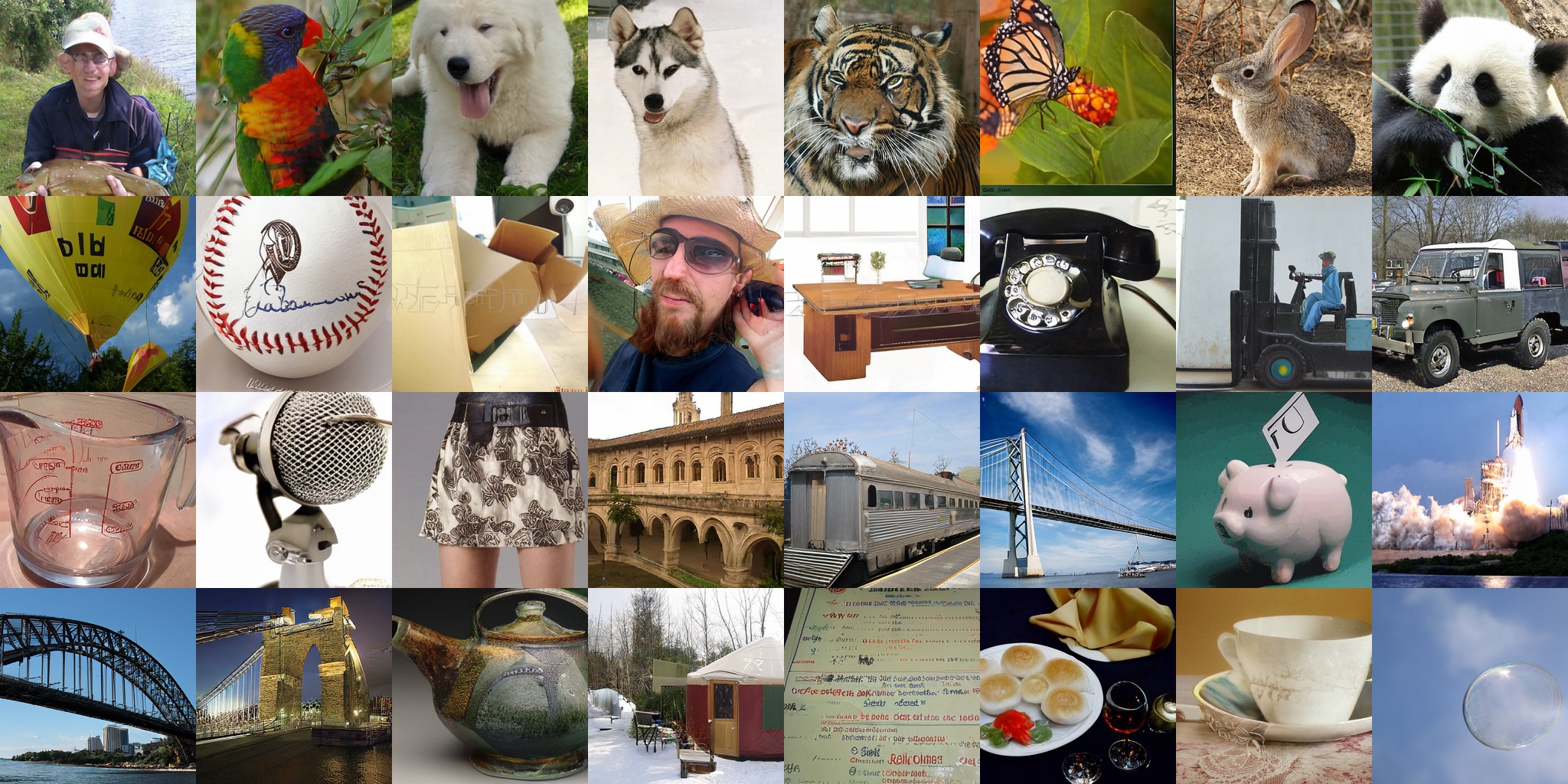}\\
    \\
    \makecell{VAR-d30 + DDO\\(FID 1.73)}&\includegraphics[width=\linewidth,align=c]{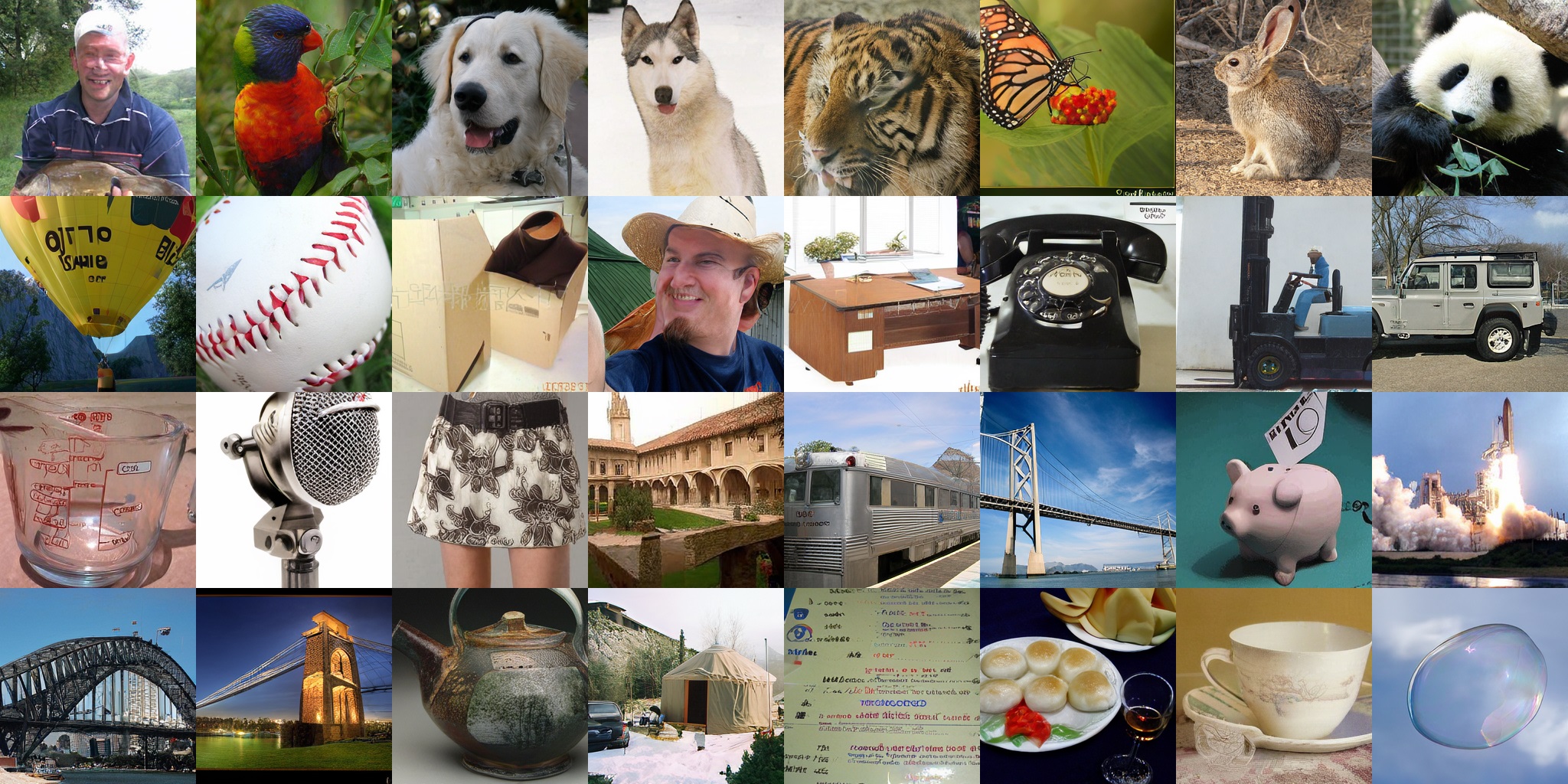}
    \end{tabular}
    }
    \caption{CFG-enhanced samples by pretrained and finetuned VAR-d30.}
    \vspace{-0.1in}
\end{figure}


\end{document}